\documentclass[acmlarge,screen,nonacm]{acmart}

\acmJournal{JOCCH}

\acmYear{2026}
\copyrightyear{2026}
\acmDOI{10.1145/3846195}
\setcopyright{acmlicensed}

\usepackage{booktabs}
\usepackage{multirow}
\usepackage{graphicx}

\graphicspath{{images/}}

\begin{document}

\title{TEXMET: A Deep Learning Benchmark for Archaeological Textile Reconstruction}

\author{Muhammad Hamza Zafar}
\orcid{0009-0006-3813-361X}
\email{hamzafer3@gmail.com}
\affiliation{%
  \institution{Norwegian University of Science and Technology (NTNU)}
  \department{Department of Computer Science, The Norwegian Colour and Visual Computing Laboratory (Colourlab)}
  \city{Gj\o{}vik}
  \country{Norway}
}

\author{Davit Gigilashvili}
\orcid{0000-0002-6956-6569}
\email{davit.gigilashvili@ntnu.no}
\affiliation{%
  \institution{Norwegian University of Science and Technology (NTNU)}
  \department{Department of Computer Science, The Norwegian Colour and Visual Computing Laboratory (Colourlab)}
  \city{Gj\o{}vik}
  \country{Norway}
}

\author{Jon Yngve Hardeberg}
\orcid{0000-0003-1150-2498}
\email{jon.hardeberg@ntnu.no}
\affiliation{%
  \institution{Norwegian University of Science and Technology (NTNU)}
  \department{Department of Computer Science, The Norwegian Colour and Visual Computing Laboratory (Colourlab)}
  \city{Gj\o{}vik}
  \country{Norway}
}

\renewcommand{\shortauthors}{Zafar et al.}

\begin{abstract}
The reconstruction of fragmented historical textiles presents a significant challenge in computational archaeology. The Oseberg ship burial, dating to 834 CE, yielded hundreds of degraded textile fragments that have resisted systematic computational analysis. While deep learning has revolutionized image-based puzzle solving, progress on archaeological textiles has been limited by the lack of domain-specific datasets and systematic evaluation of modern architectures. To address this gap, we present TEXMET, a curated dataset of 18,644 historical textile images from the Metropolitan Museum of Art's Open Access collection, representing the first textile-specific benchmark for puzzle-solving research. Using TEXMET, we conduct a systematic evaluation of three state-of-the-art puzzle-solving models: JPDVT, FCViT, and DiffAssemble. Our experiments reveal that the transformer-based models pre-trained on standard datasets fail to generalize to textiles, but domain-specific fine-tuning proves highly effective. The graph-based DiffAssemble model, when fine-tuned on TEXMET, achieves 85.89\% puzzle accuracy on textile data. Most significantly, applying this fine-tuned model to 45 highly degraded Oseberg fragments yields a performance increase from 26.67\% to 64.44\% puzzle accuracy. These results establish TEXMET as a new benchmark for textile reconstruction and demonstrate that domain adaptation is essential for applying puzzle-solving methods to imagery of real archaeological materials.

\end{abstract}

\ccsdesc[500]{Computing methodologies~Computer vision}
\ccsdesc[500]{Computing methodologies~Neural networks}
\ccsdesc[300]{Human-centered computing~HCI theory, concepts and models}

\keywords{archaeological reconstruction, deep learning, textile fragments, puzzle solving, domain adaptation, cultural heritage, vision transformers, diffusion models, graph neural networks}

\maketitle

{\renewcommand{\thefootnote}{}\footnotetext{Accepted version of the article published in \textit{ACM Journal on Computing and Cultural Heritage} (JOCCH), 2026. The version of record is available at \url{https://doi.org/10.1145/3846195}. Dataset: \url{https://huggingface.co/datasets/hzafar/TEXMET}.}}

\section{Introduction}
\label{sec:introduction}

\begin{figure}[t]
    \centering
    \includegraphics[width=\columnwidth]{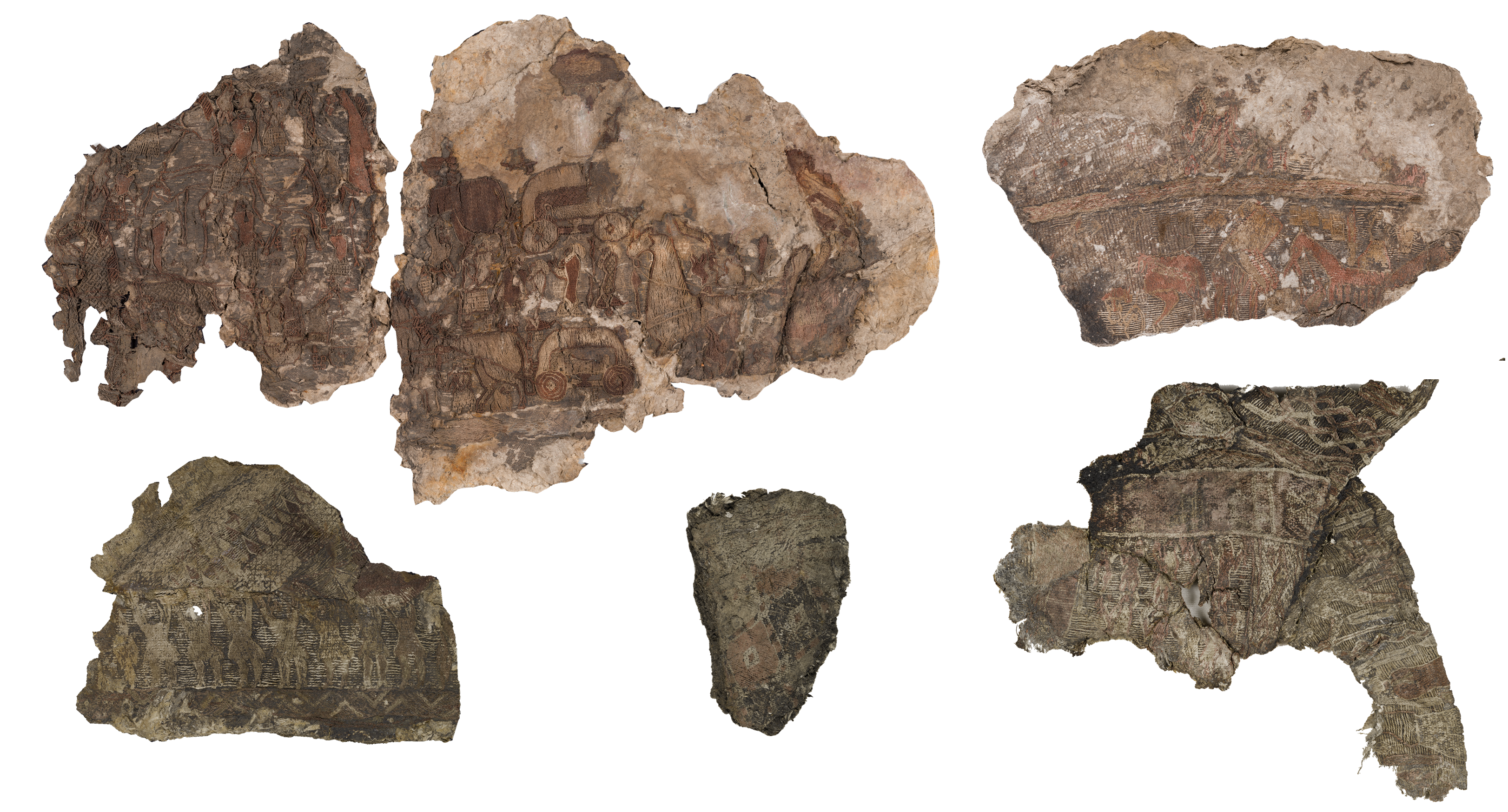}
    \caption{Examples of textile fragments from the Oseberg ship burial (834 CE). These degraded fragments present significant challenges for computational reconstruction. The fragments do not correspond to their true scale. Images by George Alexis Pantos, Museum of Cultural History, University of Oslo.}
    \label{fig:oseberg_fragments}
    \Description{Photographs of six archaeological textile fragments from the Oseberg ship burial, arranged on a black background. Each fragment is irregular in outline, torn at the edges and heavily faded, with woven figures, animals and geometric borders still partly visible in brown, gray and reddish tones.}
\end{figure}

The reconstruction of fragmented cultural heritage artifacts represents one of the most challenging tasks in modern archaeology. Among the most fascinating discoveries in Viking Age archaeology are the textile fragments found within the Oseberg ship burial, discovered in 1903 near T{\o}nsberg, Norway. This 9th-century Viking vessel, serving as the final resting place for two women of apparent high status, contained an extraordinary collection of artifacts that have provided unprecedented insights into Viking craftsmanship, daily life, and cultural practices~\cite{Vedeler2019}. The textile fragments recovered from the Oseberg burial are particularly significant, representing some of the finest examples of Viking Age weaving techniques and artistic expression~\cite{Vedeler2019,Strand2020}. However, these precious artifacts present a major challenge: they exist as hundreds of fragmented pieces, scattered and deteriorated by centuries of burial conditions. The fragments are now housed at the Museum of Cultural History in Oslo, where ongoing efforts seek to understand their original configurations (Figure~\ref{fig:oseberg_fragments}).

Recent breakthroughs in deep learning, particularly in computer vision, have created unprecedented opportunities for addressing complex pattern recognition and reconstruction challenges. Transformer architectures~\cite{Vaswani2017,ViT2021} have revolutionized image analysis tasks, with their attention mechanisms proving particularly effective for understanding spatial relationships and pattern recognition. State-of-the-art models such as JPDVT (Jigsaw Puzzles with Diffusion Vision Transformers)~\cite{JPDVT2024}, FCViT (Fragment's Coordinate Vision Transformer)~\cite{FCViT2025}, and graph diffusion-based approaches like DiffAssemble~\cite{DiffAssemble2024} have achieved impressive performance on standardized puzzle-solving benchmarks. These advances suggest that the fundamental technologies needed for automated fragment reassembly are becoming mature. However, these models have been primarily tested on clean, synthetic puzzles rather than real archaeological materials.

A significant gap exists between the controlled conditions of standard puzzle-solving benchmarks and the real-world challenges of archaeological textile reconstruction. Standard datasets typically feature clean, regularly shaped pieces arranged in regular grids (e.g., 3$\times$3, 9$\times$9), with perfect edges, uniform lighting conditions, and complete coverage of the original image~\cite{Paumard2018}. Archaeological textile fragments, by contrast, are irregularly shaped, suffer from various forms of degradation, may have missing portions, and often exist in incomplete sets where not all pieces of the original textile have been preserved~\cite{Derech2021}. Furthermore, no large-scale textile-specific dataset exists for training and evaluating puzzle-solving models, and no systematic study has evaluated modern deep learning approaches on archaeological textile data.

To address these gaps, we present TEXMET, a new dataset of 18,644 curated historical textile images from the Metropolitan Museum of Art's Open Access collection~\cite{MET_OpenAccess_2024}. To the best of our knowledge, TEXMET is the first textile-specific benchmark dataset for puzzle-solving research. Using this dataset, we conduct the first systematic evaluation of three state-of-the-art puzzle-solving models on textile data. Our experiments reveal that the transformer-based models pre-trained on standard datasets fail to generalize to textiles, but that domain-specific fine-tuning dramatically improves performance. The graph-based DiffAssemble model, when fine-tuned on TEXMET, achieves 85.89\% puzzle accuracy on textile data and, most significantly, improves from 26.67\% to 64.44\% accuracy on the highly degraded Oseberg fragments.

In summary, this work contributes TEXMET, the first textile-specific benchmark dataset for puzzle-solving research; provides the first systematic evaluation of three state-of-the-art puzzle-solving architectures on textile data and real archaeological fragments; demonstrates that domain-specific fine-tuning is essential for adapting puzzle-solving models to archaeological textile imagery; and releases all code, model checkpoints, and data to support reproducibility. We position this work as a benchmarking and domain adaptation study, a controlled first step toward computational reconstruction of archaeological textiles, rather than a complete reconstruction solution.

\section{Related Work}
\label{sec:related-work}

\subsection{Computational Puzzle Solving}

The computational jigsaw-puzzle problem has evolved significantly over the past six decades. Foundational work by Freeman and Garder~\cite{Freeman1964} introduced brute-force border matching for small puzzles using shape-only features. Subsequent decades focused on heuristic optimizations, using global shape-based reoptimization~\cite{Goldberg2002}, Markov random fields~\cite{Cho_CVPR_2010}, genetic algorithms~\cite{Sholomon2013}, and greedy color compatibility~\cite{Pomeranz_CVPR_2011}. These classical methods excelled with undamaged, uniformly shaped puzzle pieces but struggled significantly with irregular or damaged fragments.

The rise of deep learning in 2015 sparked renewed interest in puzzle-solving as a self-supervised learning task. Doersch et al.~\cite{Doersch2015} and Noroozi and Favaro~\cite{Noroozi2016} demonstrated convolutional neural network (CNN) architectures' capability to learn general visual representations through puzzle-solving pretext tasks. The introduction of transformer architectures marked a paradigm shift, with vision transformers enabling spatial reasoning directly within self-attention mechanisms~\cite{Chen2023}.

Recent breakthroughs have transformed puzzle-solving through diffusion models and graph-based approaches. JPDVT~\cite{JPDVT2024} approached puzzle-solving as a denoising problem, effectively handling missing or uncertain fragments. FCViT~\cite{FCViT2025} introduced coordinate regression for direct position prediction, claiming superior performance with a simpler architecture. DiffAssemble~\cite{DiffAssemble2024} combines graph neural networks with diffusion processes, enabling robust handling of rotation and multiple puzzle sizes within a single model. These three models represent the current state-of-the-art and form the basis of our evaluation.

\subsection{Heritage-Focused Computational Methods}

Several research efforts have applied computational puzzle-solving to cultural heritage applications. Paumard et al.~\cite{Paumard2018shortest} combined Siamese CNN-based edge matching with shortest-path algorithms, demonstrating success on a dedicated reassembly dataset of artwork images from the Metropolitan Museum of Art. Derech et al.~\cite{Derech2021} proposed fragment-extension methods specifically addressing erosion in archaeological materials, reassembling eroded artifacts and frescoes including objects from museum collections. Funkhouser et al.~\cite{Funkhouser2011} addressed the matching of fresco fragments from the Bronze Age settlement of Akrotiri (Thera) and the Roman site of Tongeren.

Despite these advances, archaeological textiles have remained notably underserved by computational puzzle-solving research. Textile fragments present unique challenges including complex weaving patterns, severe degradation from burial conditions, irregular shapes from damage, and color fading that hides visual features. The absence of systematic evaluation of modern deep learning methods on textile data represents a significant gap that this work addresses.

Several previous works explored the open archaeological puzzle problem of Oseberg textiles. Gigilashvili et al.~\cite{Gigilashvili2023archiving,Gigilashvili2024Textile} presented the first attempt to apply machine learning to the Oseberg fragment matching problem using clustering based on deep features and classical texture descriptors. Gulbrandsen~\cite{Gulbrandsen2023} developed the Artifact Assembly software tool for interactive puzzle reassembly of textile fragments. More recently, Berels~\cite{berels2025deep} investigated deep feature extraction under irregular masking conditions for content-based image retrieval in archaeological puzzle solving. A comprehensive review of computational techniques for virtual reconstruction of fragmented archaeological textiles can be found in~\cite{Gigilashvili2023heritage}. Our work builds on these efforts by introducing the first large-scale textile benchmark dataset and systematically evaluating state-of-the-art deep learning puzzle-solving models on this domain.

\subsection{Existing Datasets}

Standard puzzle-solving benchmarks have relied on natural image datasets such as ImageNet~\cite{ImageNet2009} and CelebA-HQ~\cite{CelebAHQ2018}. These datasets feature clean, high-resolution images that are synthetically fragmented into regular grid patterns. JPwLEG-3~\cite{JPwLEG3} introduced puzzles with gaps, simulating missing pieces, but still operates on clean natural images.

For cultural heritage applications, the RePAIR dataset~\cite{Tsesmelis2024} provides fragmented fresco images for benchmarking, but does not capture the unique visual properties of woven textile materials. TextileNet~\cite{TextileNet2020} offers a large collection of modern fabric images for texture classification, but these are intact images of contemporary materials unsuitable for training puzzle-solving models on historical textiles.

A critical limitation in the field has been the absence of a large-scale, publicly available dataset specifically designed for historical textile reconstruction. This gap has significantly hindered the ability to train and evaluate deep learning models on textile-specific challenges. The creation of TEXMET directly addresses this limitation, providing the first textile-specific benchmark for puzzle-solving research.

\section{The TEXMET Dataset}
\label{sec:texmet-dataset}

A critical limitation in computational textile reconstruction has been the absence of a large-scale, publicly available dataset specifically designed for historical textiles. Existing datasets are unsuitable for this domain: TextileNet~\cite{TextileNet2020} contains modern fabric swatches without fragmentation, RePAIR~\cite{Tsesmelis2024} provides fresco fragments with visual characteristics different from woven textiles, and standard puzzle benchmarks use natural images with no textile content. To address this gap, we introduce TEXMET, the first textile-specific benchmark dataset for puzzle-solving research.

\subsection{Data Collection}

TEXMET was constructed from the Metropolitan Museum of Art's Open Access collection~\cite{MET_OpenAccess_2024}, which provides high-resolution historical textile images under a CC0 public domain license. Using the MET REST API, we systematically collected textile and tapestry objects through targeted queries. The collection process involved multi-stage API harvesting to handle rate limiting and ensure completeness.

Table~\ref{tab:texmet-collection} summarizes the data collection results. From 33,588 queried objects, 27,373 contained images, yielding 55,007 total images downloaded with a 99.95\% success rate. This systematic collection required approximately three days of continuous harvesting.

\begin{table}[t]
\centering
\caption{TEXMET data collection summary from the MET archive.}
\label{tab:texmet-collection}
\begin{tabular}{@{}lr@{}}
\toprule
\textbf{Metric} & \textbf{Count} \\
\midrule
Objects queried & 33,588 \\
Objects with images & 27,373 \\
Primary images downloaded & 27,360 \\
Additional images downloaded & 27,647 \\
Total images & 55,007 \\
Download success rate & 99.95\% \\
\bottomrule
\end{tabular}
\end{table}

\subsection{Curation Pipeline}

The raw collection required extensive curation to ensure quality and relevance for puzzle-solving tasks. We employed an interactive curation pipeline using FiftyOne~\cite{moore2020fiftyone} with the following components:

\textbf{CLIP Embeddings:} GPU-accelerated CLIP ViT-B/32 embeddings~\cite{radford2021clip} were computed for all 27,360 primary images, producing 512-dimensional feature vectors for each image.

\begin{figure}[t]
    \centering
    \includegraphics[width=\columnwidth]{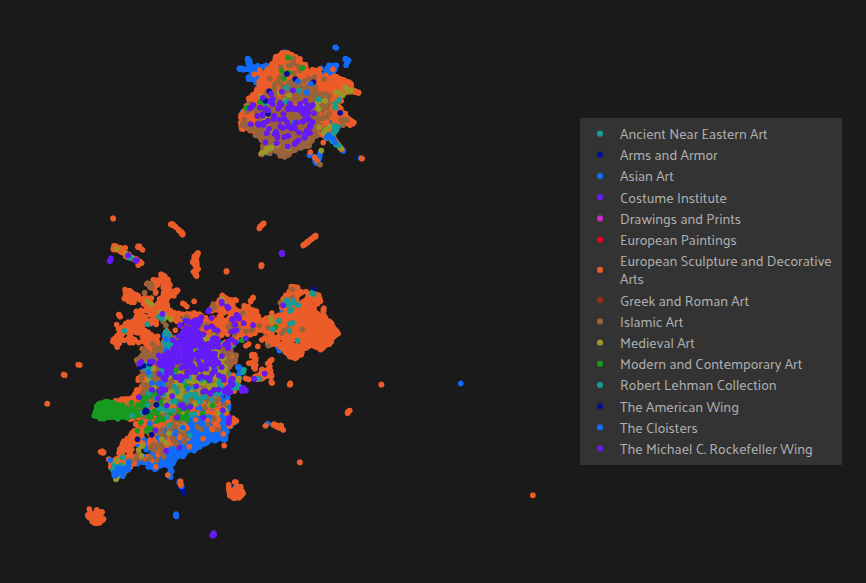}
    \caption{UMAP projection of CLIP embeddings for the textile collection, colored by MET curatorial department, as displayed in the FiftyOne curation interface. Natural clusters emerge, enabling systematic identification of outliers and quality issues.}
    \label{fig:umap-clustering}
    \Description{Scatter plot of a two-dimensional UMAP projection of CLIP image embeddings on a dark background, where each point is one textile image colored by MET curatorial department. The points form one large dense cluster and one smaller separate cluster with scattered outliers around them. A legend lists fifteen departments, including Asian Art, Costume Institute, Islamic Art, Medieval Art and The Cloisters.}
\end{figure}

\textbf{UMAP Clustering:} Embeddings were projected to three dimensions using UMAP~\cite{mcinnes2018umap}, revealing natural clusters and outliers for systematic quality assessment. Figure~\ref{fig:umap-clustering} shows the UMAP projection of CLIP embeddings, revealing natural clusters that guided systematic quality assessment.

\textbf{Similarity Search:} LanceDB~\cite{LanceDB} indexing enabled sub-second nearest-neighbor queries, allowing efficient identification of problematic image clusters.

\textbf{Quality Assessment:} Samples were classified as ``poor quality'' (irrelevant content such as armor or weapons misclassified as textiles) or ``noise'' (insufficient textile coverage, individual threads, or minimal textile elements).

\begin{figure}[t]
    \centering
    \includegraphics[width=\columnwidth]{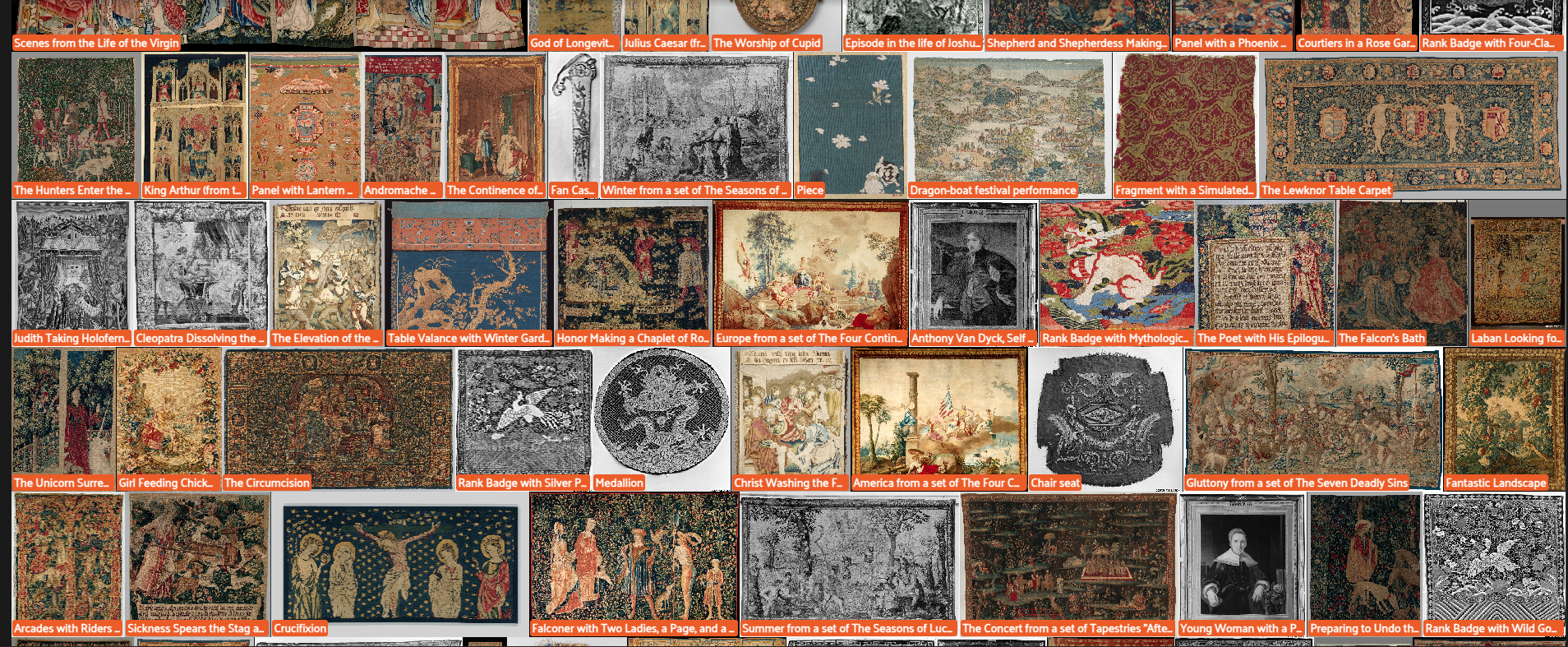}
    \caption{Overview of the final curated TEXMET dataset showing 18,644 high-quality textile images spanning diverse historical periods and cultural traditions.}
    \label{fig:texmet-grid}
    \Description{A dense grid of thumbnail images from the curated TEXMET dataset, each labeled with its object title. The thumbnails show tapestries, woven panels, rank badges, embroidered hangings and textile fragments from many periods and cultures, in both color and black and white.}
\end{figure}

The curation process removed 8,716 samples in two categories, 4,908 tagged as poor quality and 3,808 as noise, of which 2,567 were duplicate images, yielding a final dataset of 18,644 high-quality textile images representing 68.1\% of the original collection.
Figure~\ref{fig:texmet-grid} provides an overview of the final curated dataset.

\subsection{Dataset Statistics}

Table~\ref{tab:texmet-stats} presents the technical characteristics of the final TEXMET dataset. The collection demonstrates consistent quality with an average resolution of 3.3 megapixels and dimensions suitable for puzzle-solving experiments.

\begin{table}[t]
\centering
\caption{TEXMET dataset statistics.}
\label{tab:texmet-stats}
\begin{tabular}{@{}lr@{}}
\toprule
\textbf{Property} & \textbf{Value} \\
\midrule
Total images & 18,644 \\
Train / Val / Test split & 14,915 / 1,864 / 1,865 \\
Average dimensions & 1,557 $\times$ 1,631 pixels \\
Average resolution & 3.3 MP \\
Color images (RGB) & 92.0\% \\
Black and white images & 8.0\% \\
Total dataset size & 26.5 GB \\
\bottomrule
\end{tabular}
\end{table}

The dataset maintains strong representation across 14 MET curatorial departments, preserving global diversity in textile traditions. Table~\ref{tab:texmet-departments} shows the department distribution, with European Art contributing the largest proportion followed by Asian Art and Islamic Art.

\begin{table}[t]
\centering
\caption{Department distribution in TEXMET.}
\label{tab:texmet-departments}
\begin{tabular}{@{}lrr@{}}
\toprule
\textbf{Department} & \textbf{Samples} & \textbf{\%} \\
\midrule
European Sculpture and Decorative Arts & 8,516 & 45.7 \\
Asian Art & 5,229 & 28.0 \\
Islamic Art & 3,029 & 16.2 \\
Other departments (11) & 1,870 & 10.0 \\
\midrule
\textbf{Total} & \textbf{18,644} & \textbf{100.0} \\
\bottomrule
\end{tabular}
\end{table}

The supplementary material presents the technical characteristics of TEXMET images in detail, showing consistent quality suitable for puzzle-solving experiments.

\subsection{Comparison with Existing Datasets}

Table~\ref{tab:dataset-comparison} compares TEXMET with existing datasets used in puzzle-solving and textile research. TEXMET uniquely combines historical textile content, open licensing, and suitability for puzzle-solving evaluation.

\begin{table}[t]
\centering
\caption{Comparison of TEXMET with related datasets.}
\label{tab:dataset-comparison}
\begin{tabular}{@{}lcccc@{}}
\toprule
\textbf{Dataset} & \textbf{Domain} & \textbf{Size} & \textbf{Fragmented} & \textbf{License} \\
\midrule
ImageNet & Natural images & 1.28M & No & Research \\
CelebA-HQ & Faces & 30k & No & Research \\
TextileNet & Modern fabrics & 760k & No & Academic \\
RePAIR & Frescoes & 1k+ & Yes & Restricted \\
\textbf{TEXMET} & \textbf{Historical textiles} & \textbf{18,644} & \textbf{Simulated} & \textbf{CC0} \\
\bottomrule
\end{tabular}
\end{table}

To the best of our knowledge, TEXMET represents the first comprehensively curated textile dataset specifically designed for computational puzzle-solving research. The dataset, along with all code and trained models, will be made publicly available.

\section{Methods}
\label{sec:methods}

This section describes the three state-of-the-art puzzle-solving models evaluated in this research: JPDVT, FCViT, and DiffAssemble. Each represents a different architectural approach to the puzzle-solving problem.

\subsection{Problem Formulation}

Given an image divided into $N$ non-overlapping fragments $\{x_i\}_{i=1}^{N}$, the puzzle-solving task aims to determine the correct spatial arrangement that reconstructs the original image. Formally, for each fragment $x_i$, the model must predict its correct grid position $l_i$. The reconstructed image is then:
\begin{equation}
X = \bigcup_{i=1}^{N}(x_i @ l_i),
\end{equation}
where $x_i @ l_i$ denotes fragment $x_i$ placed at location $l_i$.

\subsection{JPDVT: Diffusion Vision Transformer}

\begin{figure}[t]
    \centering
    \includegraphics[width=\columnwidth]{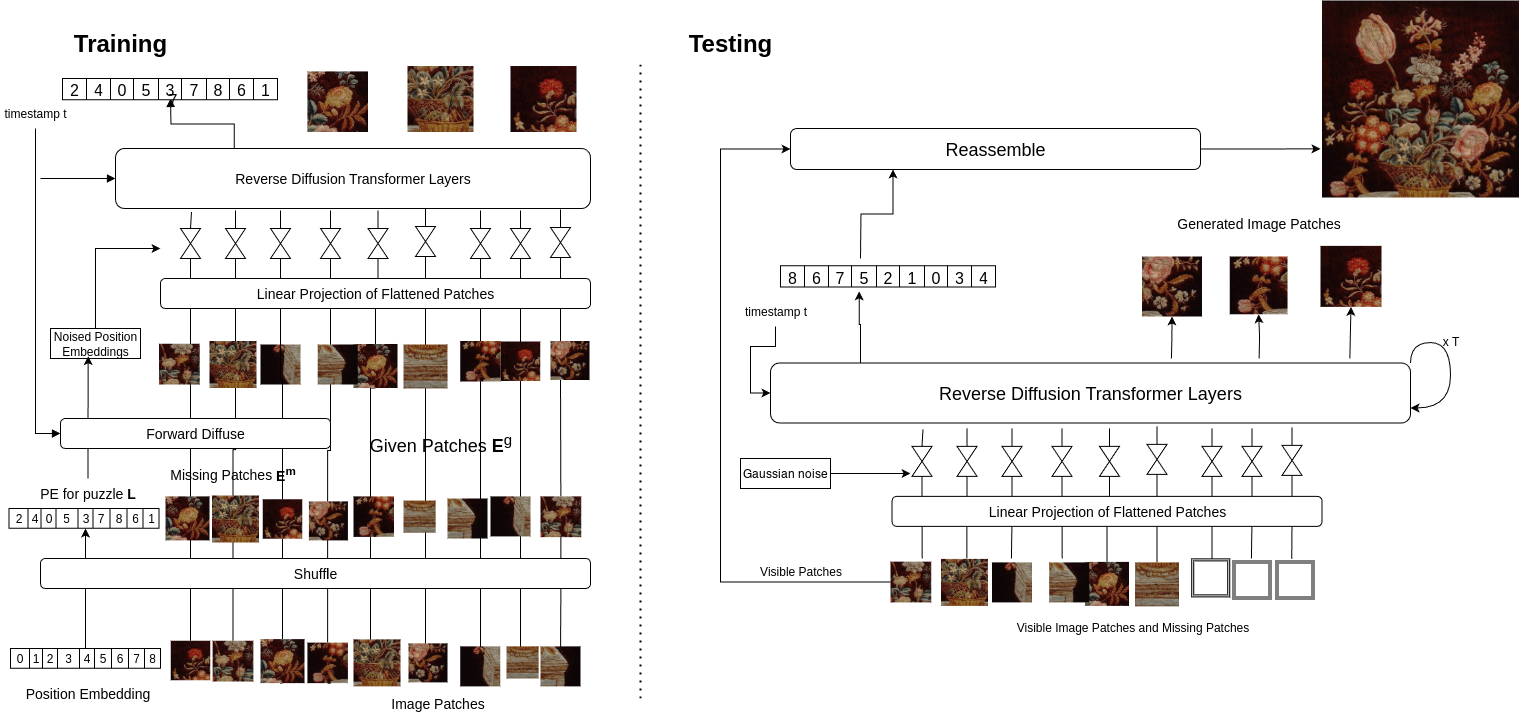}
    \caption{JPDVT architecture showing the Vision Transformer backbone integrated with the diffusion process for position prediction. Figure by the authors, created using Canva.}
    \label{fig:jpdvt-arch}
    \Description{Two-part pipeline diagram of JPDVT. On the left, the training path: image patches are shuffled, their position embeddings are noised by forward diffusion, and the patches pass through a linear projection of flattened patches and reverse diffusion transformer layers that output the predicted patch ordering. On the right, the testing path: visible patches, missing patches and Gaussian noise pass through the same transformer layers for T iterations, generating the missing patches and reassembling the complete textile image.}
\end{figure}

JPDVT~\cite{JPDVT2024} approaches puzzle-solving as a conditional diffusion problem~\cite{DDPM2020}. The model uses a Vision Transformer (ViT) encoder to extract visual features from each fragment, then employs a diffusion process to predict positional encodings. Figure~\ref{fig:jpdvt-arch} illustrates the architecture.

During training, Gaussian noise is progressively added to the positional encodings (forward diffusion). During inference, the model learns to reverse this process, denoising the positional information conditioned on visual features. This approach naturally handles missing pieces through mask token embeddings, making it robust to incomplete puzzles.

The authors reported 68.7\% puzzle accuracy and 83.3\% piece accuracy on ImageNet 3$\times$3 puzzles. A key advantage of JPDVT is its ability to model long-range dependencies through the transformer architecture and handle missing fragments through the diffusion framework.

\subsection{FCViT: Coordinate Regression}

\begin{figure}[t]
    \centering
    \includegraphics[width=\columnwidth]{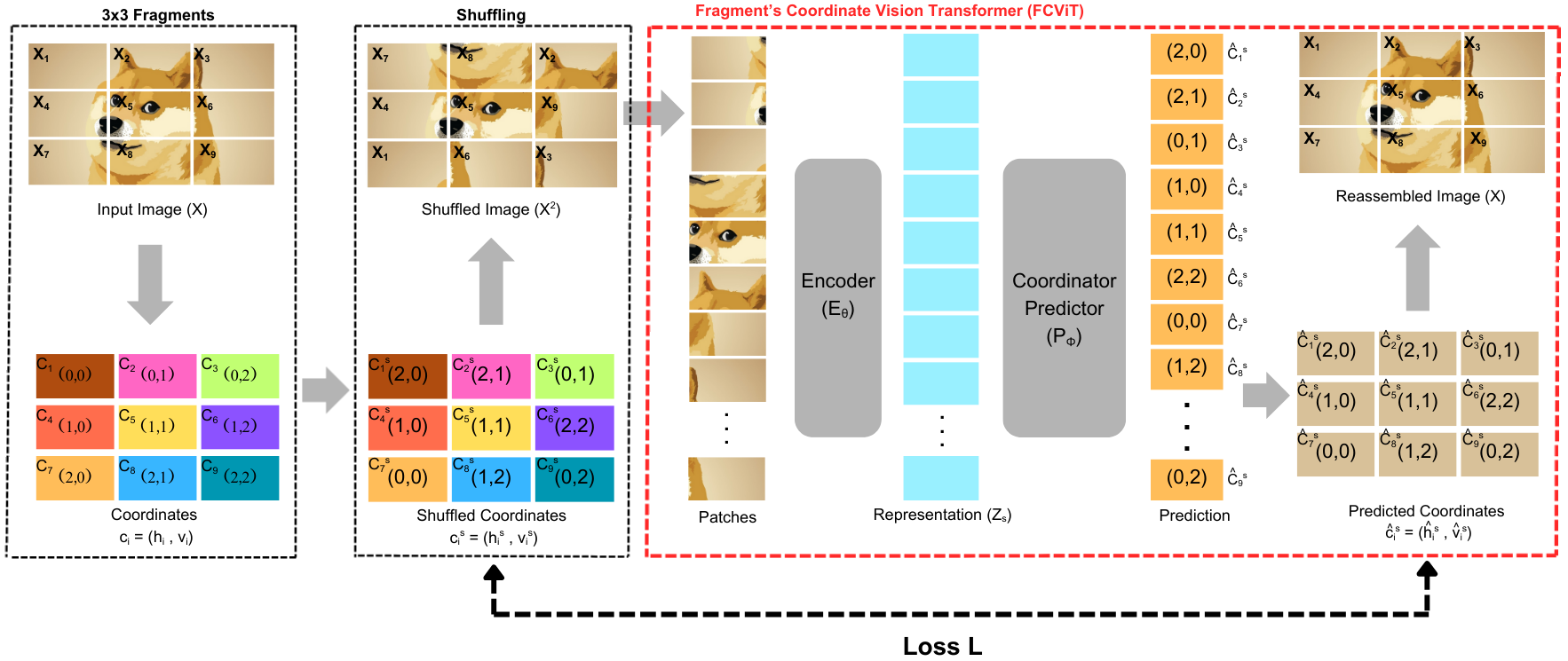}
    \caption{FCViT architecture illustrating the coordinate regression process where fragment positions are directly predicted. Figure by the authors, created using Canva.}
    \label{fig:fcvit-arch}
    \Description{Flow diagram of FCViT. An input image is divided into a 3 by 3 grid of fragments with ground-truth coordinates, the fragments are shuffled, and the shuffled patches pass through an encoder and a coordinate predictor that outputs a predicted row and column coordinate for each fragment. The predicted coordinates are used to reassemble the image, and a loss term compares predicted coordinates with true coordinates.}
\end{figure}

FCViT~\cite{FCViT2025} reformulates puzzle-solving as a coordinate regression task rather than classification. Instead of predicting permutations, the model directly predicts horizontal and vertical coordinates for each fragment using a Vision Transformer encoder followed by a multilayer perceptron (MLP) prediction head. Figure~\ref{fig:fcvit-arch} illustrates the architecture.

The training objective minimizes a smooth $L_1$ loss between predicted and ground-truth coordinates:
\begin{equation}
\mathcal{L}_{coord} = \mathcal{L}_{coord}^h + \mathcal{L}_{coord}^v,
\end{equation}
where the losses for horizontal and vertical coordinates are computed separately.

FCViT achieved 78.9\% puzzle accuracy and 90.6\% piece accuracy on ImageNet 3$\times$3 puzzles, outperforming JPDVT. However, a notable limitation is a tendency toward duplicate placements, where multiple fragments may be assigned to similar coordinates due to the regression formulation lacking explicit uniqueness constraints.

\subsection{DiffAssemble: Graph-Based Diffusion}

\begin{figure}[t]
    \centering
    \includegraphics[width=\columnwidth]{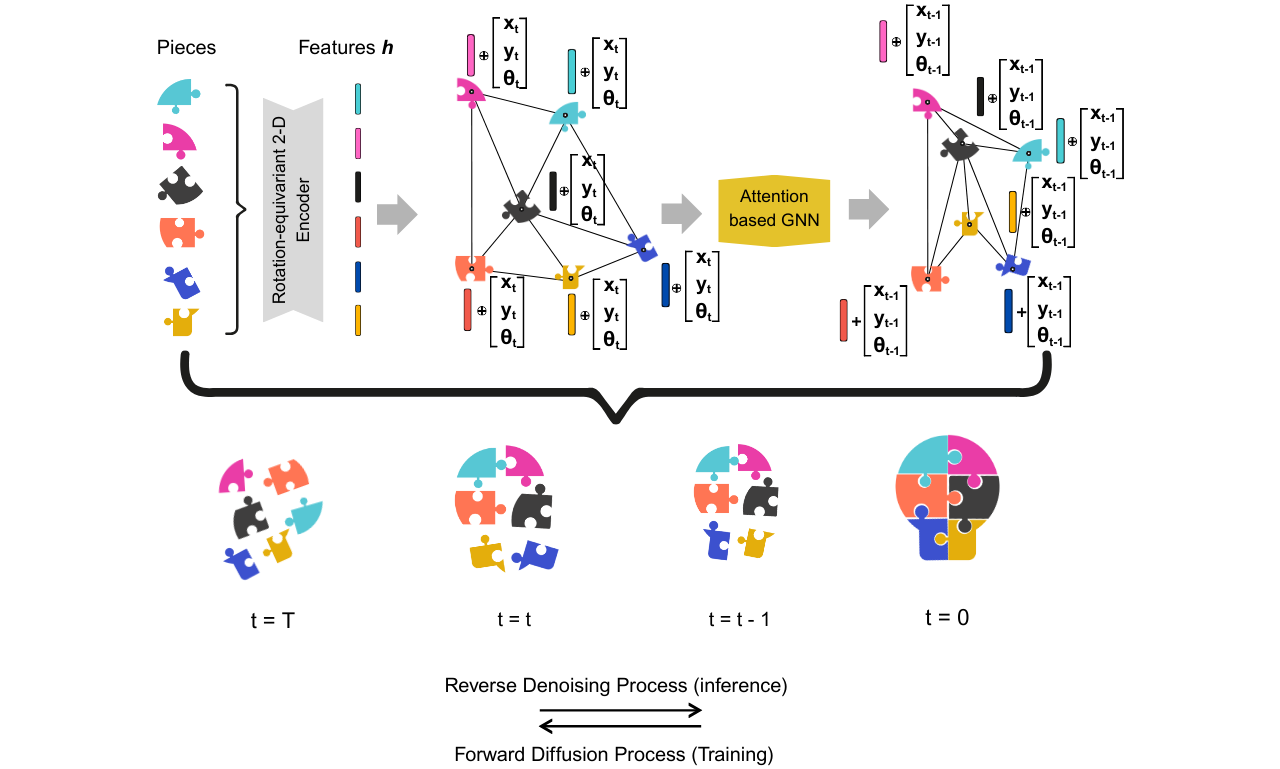}
    \caption{DiffAssemble architecture demonstrating the graph-based diffusion approach for 2D puzzle reassembly. Figure by the authors, created using Canva.}
    \label{fig:diffassemble-arch}
    \Description{Diagram of DiffAssemble. Puzzle pieces are embedded by a rotation-equivariant 2D encoder, each piece becomes a graph node combining its feature vector with position and rotation values at diffusion step t, and an attention-based graph neural network denoises the node attributes to step t minus 1. Below, four snapshots show the pieces moving from scattered at t equals T to a fully assembled image at t equals 0, with arrows marking forward diffusion during training and reverse denoising at inference.}
\end{figure}

DiffAssemble~\cite{DiffAssemble2024} represents puzzles as graphs, where each fragment is a node and edges encode spatial relationships. A graph neural network (GNN) combined with diffusion denoising predicts both translations and rotations for each piece. Figure~\ref{fig:diffassemble-arch} illustrates the architecture.

The forward diffusion adds noise to fragment positions and orientations:
\begin{equation}
q(x_t | x_0) = \mathcal{N}(x_t; \sqrt{\alpha_t} x_0, (1-\alpha_t)\mathbf{I}),
\end{equation}
where $x_0$ encodes the initial pose and $\alpha_t$ controls diffusion at timestep $t$. The reverse process iteratively refines predictions through attention-based GNN layers.

A key advantage of DiffAssemble is its unified architecture that handles multiple puzzle sizes (6$\times$6 to 12$\times$12) with a single 26.7M parameter model. The model achieved 99.51\% piece accuracy on 6$\times$6 puzzles and 97.76\% on 12$\times$12 puzzles. It also demonstrates robustness to missing pieces, maintaining 76.5\% accuracy on 12$\times$12 puzzles with 30\% missing fragments.

\subsection{Model Capabilities Comparison}

Table~\ref{tab:model-capabilities} summarizes the key capabilities of each model relevant to archaeological textile reconstruction.

\begin{table}[t]
\centering
\caption{Comparison of puzzle-solving model capabilities.}
\label{tab:model-capabilities}
\begin{tabular}{@{}lcccc@{}}
\toprule
\textbf{Model} & \textbf{Rotation} & \textbf{Missing Pieces} & \textbf{Multi-size} & \textbf{Approach} \\
\midrule
JPDVT & No & Yes & No & Diffusion \\
FCViT & No & No & No & Regression \\
DiffAssemble & Yes & Yes & Yes & Graph-Diffusion \\
\bottomrule
\end{tabular}
\end{table}

DiffAssemble offers the most comprehensive feature set for archaeological applications, supporting rotation, missing pieces, and variable puzzle sizes within a single model.

\subsection{Evaluation Metrics}

We employ two complementary metrics:

\textbf{Puzzle-Level Accuracy} measures the percentage of puzzles completely solved correctly. All pieces must be in their exact correct positions for a puzzle to count as solved:
\begin{equation}
\text{Puzzle Accuracy} = \frac{\text{Perfectly solved puzzles}}{\text{Total puzzles}} \times 100\%
\end{equation}

\textbf{Piece-Level Accuracy} (Fragment Accuracy) measures the percentage of individual pieces correctly positioned across all puzzles, providing insight into partial reconstruction success:
\begin{equation}
\text{Piece Accuracy} = \frac{\text{Correctly placed pieces}}{\text{Total pieces}} \times 100\%
\end{equation}

These metrics together capture both exact reassembly of all fragments (puzzle accuracy) and partial success in challenging cases (piece accuracy).

\section{Experimental Setup}
\label{sec:experimental-setup}

\subsection{Datasets}

We evaluate the three puzzle-solving models on three datasets representing different domains:

\textbf{ImageNet-1k}~\cite{ImageNet2009} serves as the baseline training dataset, containing 1,281,167 natural images. All three models were initially trained or reproduced on ImageNet following their original protocols.

\textbf{TEXMET} (Section~\ref{sec:texmet-dataset}) provides domain-specific training data with 18,644 historical textile images split into 14,915 training, 1,864 validation, and 1,865 test images.

\textbf{Oseberg Fragments} consist of 45 textile fragments from the Oseberg ship burial (834 CE), representing highly degraded archaeological materials. These fragments exhibit an average masked area of 75.60\%, measured from segmentation masks of the fragment photographs, indicating severe degradation with minimal visual context compared to standard puzzle datasets.

\subsection{Hardware and Implementation}

All experiments were conducted on a high-performance computing cluster equipped with NVIDIA A100 and H100 GPUs. Training utilized distributed computing across multiple GPUs to accommodate the computational demands of transformer and diffusion-based architectures.

To ensure reproducibility, all experiments used a fixed random seed of 42 with PyTorch deterministic mode enabled. Training progress was monitored through systematic logging of accuracy and loss metrics.

\subsection{Training Protocol}

\textbf{Model Reproduction:} Each model was reproduced following its original training protocol. JPDVT used AdamW optimization~\cite{AdamW2019} with the learning rate schedule specified in the original paper. FCViT followed the training configuration from its authors' implementation. DiffAssemble was trained on CelebA-HQ for 158 epochs with batch size 8, using the DDIM sampling strategy~\cite{DDIM2021} and ResNet18-equivalent backbone~\cite{ResNet2016} with transformer layers, following the original paper's protocol. The ImageNet and TEXMET experiments used the implementation's default configuration: an EfficientNet-B0 backbone~\cite{EfficientNet2019} with transformer layers, batch size 128 across four GPUs, and rotation disabled.

\textbf{Fine-tuning:} For domain adaptation experiments, models pre-trained on ImageNet were fine-tuned on TEXMET. DiffAssemble fine-tuning continued for 150 additional epochs on TEXMET training data while monitoring performance on both TEXMET validation and Oseberg evaluation sets.

\subsection{Evaluation Protocol}
\label{sec:eval-protocol}

\textbf{Puzzle Generation:} Each evaluation puzzle is generated from a single source image. The image is center-cropped and resized to the model's input resolution (JPDVT: 192$\times$192; the JPDVT-T $4\times4$ variant: 256$\times$256; per-model sizes follow the original implementations), tiled into $G\times G$ equal square tiles, and the tile order is permuted with a seeded random permutation. One permutation is drawn per image per evaluation. Tile orientation is kept fixed (no rotations) in all TEXMET and Oseberg evaluations: JPDVT and FCViT do not support rotation, and rotation was disabled for DiffAssemble for comparability. All $G^2$ tiles are present and known to belong to the same puzzle; the benchmark contains no missing pieces and no distractor tiles.

\textbf{Splits and Leakage:} TEXMET is split at the image level by a seeded random shuffle (seed 42) into 80/10/10 train, validation, and test partitions (14,915 / 1,864 / 1,865 images). Each image is the primary image of a distinct museum object, so the split is simultaneously object-level: no object appears in more than one partition, and because every puzzle is constructed from exactly one image, no puzzle tiles can span partitions by construction. We note one honest caveat: distinct museum objects can originate from related historical textiles (for example, several catalogued fragments of one source fabric), and no metadata-level grouping was applied for such cases.

\textbf{Ground Truth and Metrics:} The ground truth for each puzzle is the original tile arrangement of the intact image; a tile is placed correctly when its predicted position index equals its true index. Puzzle accuracy is the fraction of test images whose entire permutation is recovered exactly, and fragment accuracy is the mean fraction of correctly placed tiles per image (Section~\ref{sec:methods}). All TEXMET results are computed over the full 1,865-image test partition unless a subset is stated explicitly (the masking and inpainting experiments use 1,000 ImageNet validation images per condition). Evaluations use a single run per configuration with seeded shuffling; where repeated runs revealed seed sensitivity, this is reported alongside the affected results (Section~\ref{sec:masking}).

\textbf{Puzzle Configurations:} JPDVT and FCViT were evaluated on 3$\times$3 puzzles (9 pieces) following their original benchmarks. DiffAssemble was evaluated on multiple sizes from 6$\times$6 to 12$\times$12, with 3$\times$3 configurations used for cross-model comparisons. Oseberg fragments were evaluated using 3$\times$3 puzzles across all models.

\textbf{Cross-Dataset Evaluation:} To assess domain generalization, models trained on ImageNet were directly evaluated on TEXMET and Oseberg without additional training, measuring the domain gap between natural images and textile data.

\textbf{Domain Adaptation:} Models were fine-tuned on TEXMET and re-evaluated on both TEXMET test data and Oseberg fragments to measure the effectiveness of domain-specific training.

\textbf{Oseberg Evaluation:} The 45 Oseberg fragment images are evaluated with the same single-image pipeline: each fragment image is individually divided into an artificial 3$\times$3 grid and reassembled in isolation. No inter-fragment ground truth exists, so this evaluation measures within-fragment reordering, not cross-fragment matching (see the discussion of this simplification in Section~\ref{sec:discussion}).

\textbf{Inscribed Cropping:} To investigate whether models learn textile patterns or merely shape information, we evaluated using inscribed square crops that remove corner shape cues while preserving central textile content (for the Oseberg fragments, 206 manually annotated square crops from the 45 fragment images).

\section{Results}
\label{sec:results}

This section presents experimental results organized by research question: model reproduction, ImageNet baseline comparison, cross-domain evaluation, Oseberg baseline performance, domain adaptation effectiveness, analysis of learned representations, degradation and inpainting effects, and an ablation analysis.

\subsection{\texorpdfstring{Model Reproduction}{Model Reproduction}}

We first verify our implementations by reproducing the benchmark results reported in each model's original publication. Table~\ref{tab:reproduction} compares our results with those reported in the original papers for JPDVT and FCViT, whose publications report ImageNet 3$\times$3 results.

\begin{table}[t]
\centering
\caption{Model reproduction results on ImageNet 3$\times$3 puzzles. Puzzle and fragment accuracy (\%) compared to original publications.}
\label{tab:reproduction}
\begin{tabular}{@{}lcccc@{}}
\toprule
\multirow{2}{*}{\textbf{Model}} & \multicolumn{2}{c}{\textbf{Original Paper}} & \multicolumn{2}{c}{\textbf{Our Reproduction}} \\
\cmidrule(lr){2-3} \cmidrule(lr){4-5}
 & Puzzle & Fragment & Puzzle & Fragment \\
\midrule
JPDVT & 68.7 & 83.3 & 67.89 & 80.02 \\
FCViT & 78.9 & 90.6 & 68.78 & 84.21 \\
\bottomrule
\end{tabular}
\end{table}

JPDVT 's puzzle accuracy was reproduced within 1 percentage point of the original result (fragment accuracy within 3.3 points). FCViT showed a 10 percentage point gap from reported performance despite following the original training protocol, representing a limitation of our implementation. The original DiffAssemble publication reports no ImageNet results, so we verify our DiffAssemble implementation on the benchmark its authors do report: CelebA-HQ across multiple puzzle sizes. DiffAssemble closely matched the original paper's results across all puzzle sizes, demonstrating robust reproducibility.

\begin{figure*}[t]
    \centering
    \includegraphics[width=0.48\textwidth]{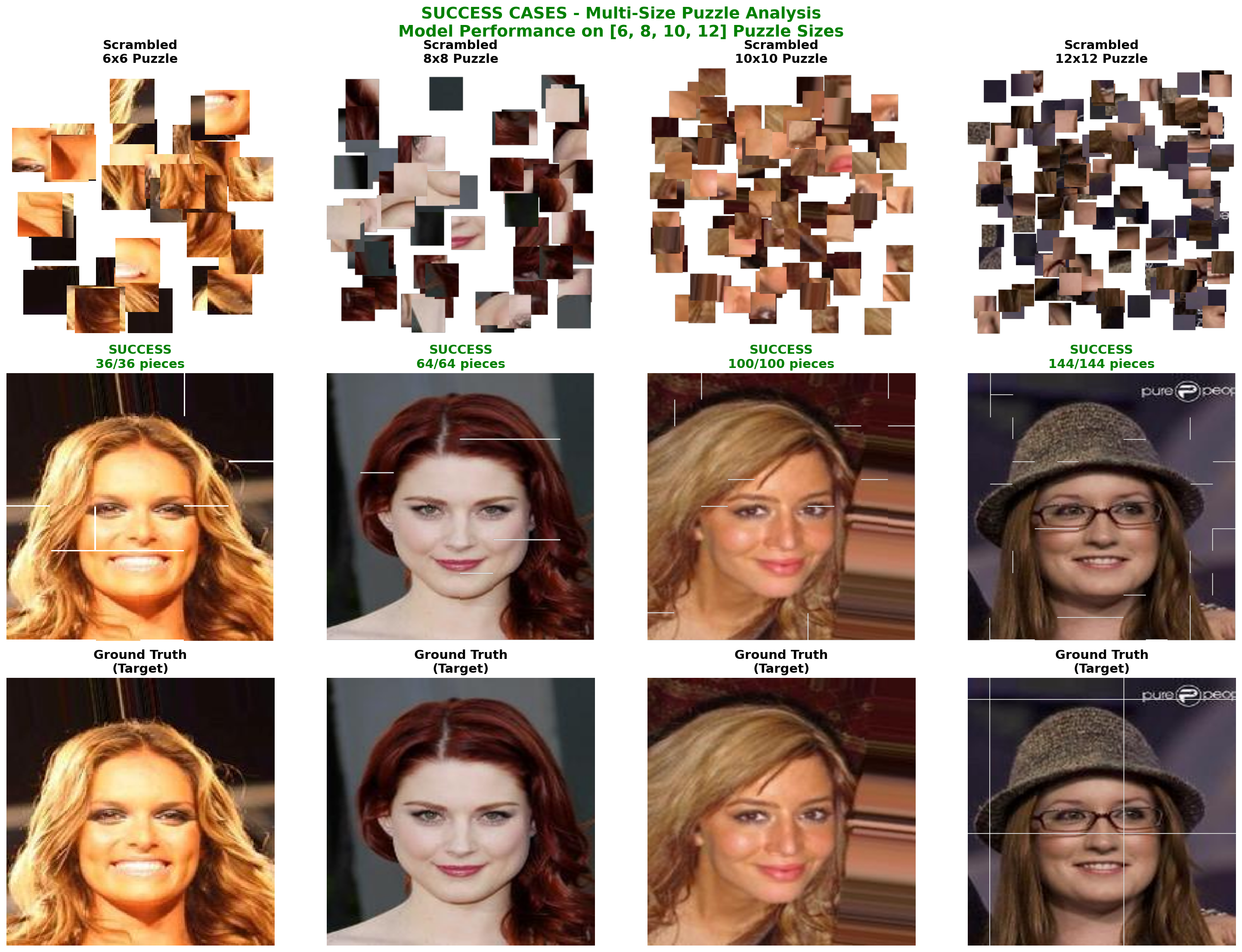}
    \includegraphics[width=0.48\textwidth]{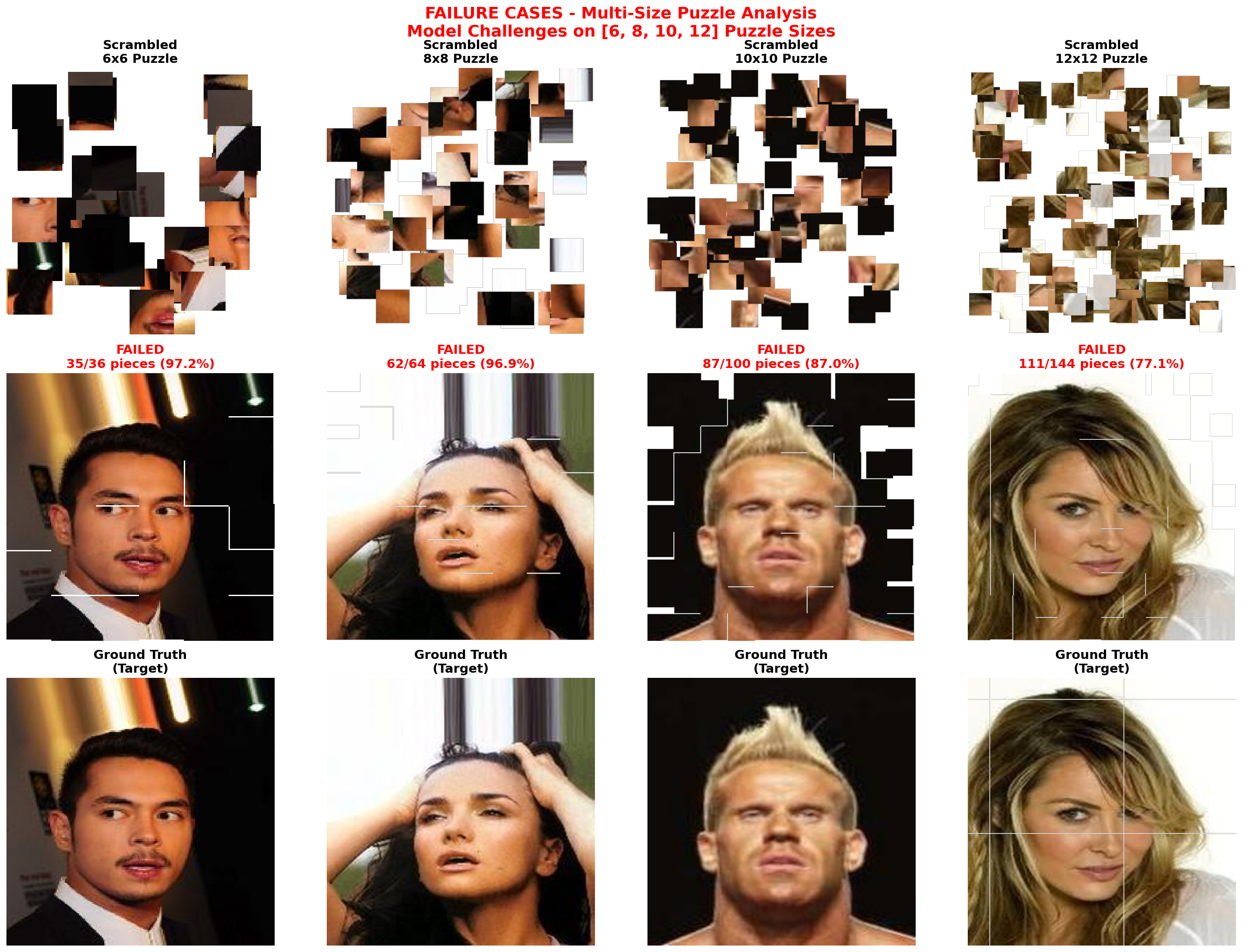}
    \caption{DiffAssemble reconstruction examples on CelebA-HQ~\cite{CelebAHQ2018} across puzzle sizes (6$\times$6 to 12$\times$12). Left: successful reconstructions showing correct piece placement. Right: failure cases revealing common error patterns.}
    \label{fig:diffassemble-gallery}
    \Description{Two side-by-side galleries of DiffAssemble reconstructions of CelebA-HQ face images at puzzle sizes 6 by 6, 8 by 8, 10 by 10 and 12 by 12. Each column shows the scrambled pieces, the model's reconstruction and the ground-truth image. On the left, success cases place every piece correctly: 36 of 36, 64 of 64, 100 of 100 and 144 of 144. On the right, failure cases degrade as the puzzle grows: 35 of 36 (97.2 percent), 62 of 64 (96.9 percent), 87 of 100 (87.0 percent) and 111 of 144 (77.1 percent).}
\end{figure*}

Table~\ref{tab:diffassemble-sizes} presents DiffAssemble performance across multiple puzzle configurations, showing graceful degradation as puzzle complexity increases. Figure~\ref{fig:diffassemble-gallery} presents qualitative examples of DiffAssemble reconstructions across multiple puzzle sizes, illustrating both successful assemblies and failure cases.

\begin{table}[t]
\centering
\caption{DiffAssemble performance across puzzle sizes on CelebA-HQ.}
\label{tab:diffassemble-sizes}
\begin{tabular}{@{}lcc@{}}
\toprule
\textbf{Puzzle Size} & \textbf{Piece Accuracy (\%)} & \textbf{Puzzle Accuracy (\%)} \\
\midrule
6$\times$6 (36 pieces) & 99.32 & 91.93 \\
8$\times$8 (64 pieces) & 98.80 & 83.39 \\
10$\times$10 (100 pieces) & 98.57 & 75.53 \\
12$\times$12 (144 pieces) & 98.05 & 68.07 \\
\bottomrule
\end{tabular}
\end{table}

\subsection{\texorpdfstring{ImageNet Baseline}{ImageNet Baseline}}
\label{sec:imagenet-baseline}

\begin{figure}[t]
    \centering
    \includegraphics[width=\columnwidth]{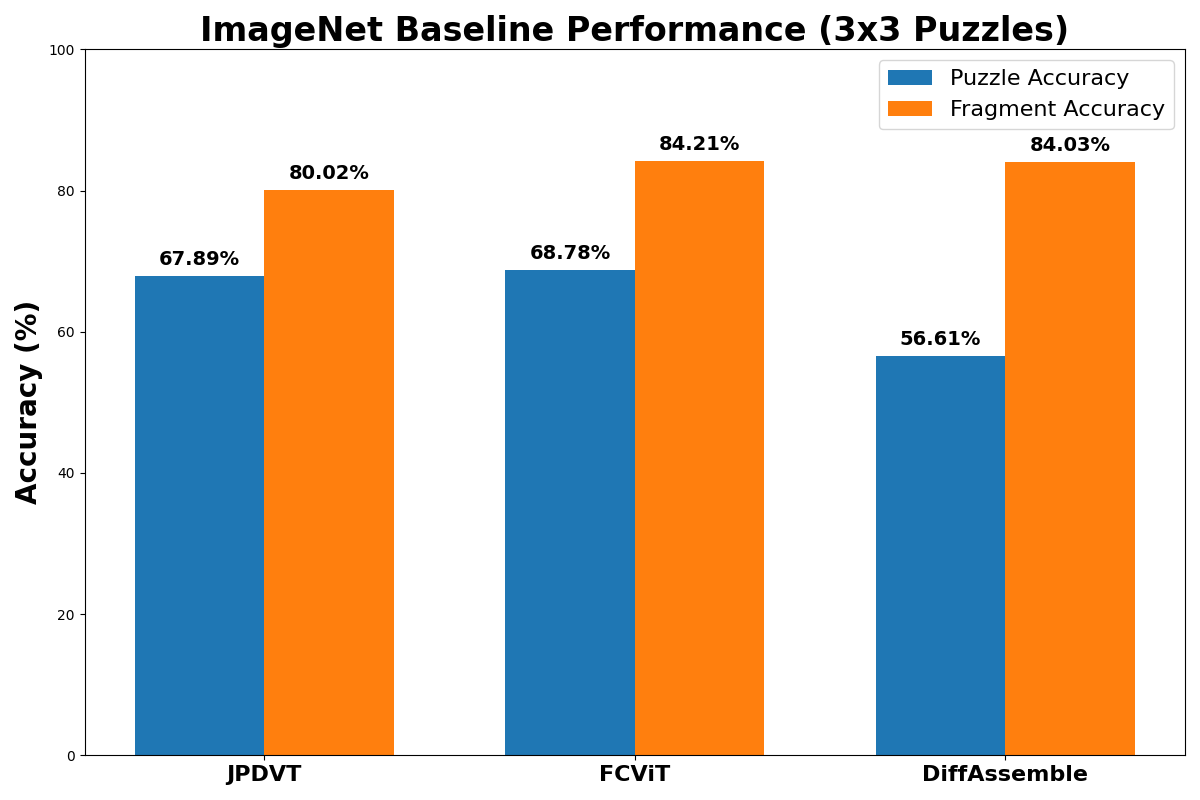}
    \caption{Baseline performance comparison of JPDVT, FCViT, and DiffAssemble on ImageNet 3$\times$3 puzzles showing puzzle and fragment accuracy.}
    \label{fig:imagenet-baseline}
    \Description{Grouped bar chart of baseline accuracy on ImageNet 3 by 3 puzzles for three models, with paired bars for puzzle accuracy and fragment accuracy. JPDVT reaches 67.89 percent puzzle accuracy and 80.02 percent fragment accuracy, FCViT 68.78 and 84.21 percent, and DiffAssemble 56.61 and 84.03 percent.}
\end{figure}

With the implementations verified, we establish a common starting point for the cross-domain experiments: all three models evaluated on ImageNet 3$\times$3 puzzles under the unified evaluation protocol of Section~\ref{sec:eval-protocol}. JPDVT reaches 67.89\% puzzle and 80.02\% fragment accuracy, FCViT 68.78\% and 84.21\%, and DiffAssemble 56.61\% and 84.03\%. DiffAssemble trails the two transformer models at the puzzle level while remaining competitive at the fragment level; this ranking reverses once the domain shifts to textiles. Figure~\ref{fig:imagenet-baseline} visualizes this comparison.

\subsection{Cross-Domain Evaluation}

To assess domain generalization, we evaluated ImageNet-trained models directly on TEXMET without any domain-specific training. This experiment reveals the domain gap between natural images and historical textiles.

\begin{table}[t]
\centering
\caption{Cross-domain evaluation: ImageNet-trained models tested on TEXMET 3$\times$3 puzzles without fine-tuning.}
\label{tab:cross-domain}
\begin{tabular}{@{}lcc@{}}
\toprule
\textbf{Model} & \textbf{Puzzle Accuracy (\%)} & \textbf{Fragment Accuracy (\%)} \\
\midrule
JPDVT & 32.92 & 55.89 \\
FCViT & 8.30 & 28.25 \\
DiffAssemble & 65.08 & 86.63 \\
\bottomrule
\end{tabular}
\end{table}

The results in Table~\ref{tab:cross-domain} reveal substantial performance variation across architectures. FCViT experienced near-complete collapse to 8.30\%, suggesting its coordinate regression approach is highly sensitive to domain-specific visual features. JPDVT maintained moderate performance at 32.92\%, indicating partial generalization capability. DiffAssemble demonstrated the strongest cross-domain robustness, maintaining 65.08\% puzzle accuracy, a level that even slightly exceeds its 56.61\% ImageNet 3$\times$3 baseline under the same evaluation protocol (Section~\ref{sec:imagenet-baseline}).

\begin{figure}[t]
    \centering
    \includegraphics[width=\columnwidth]{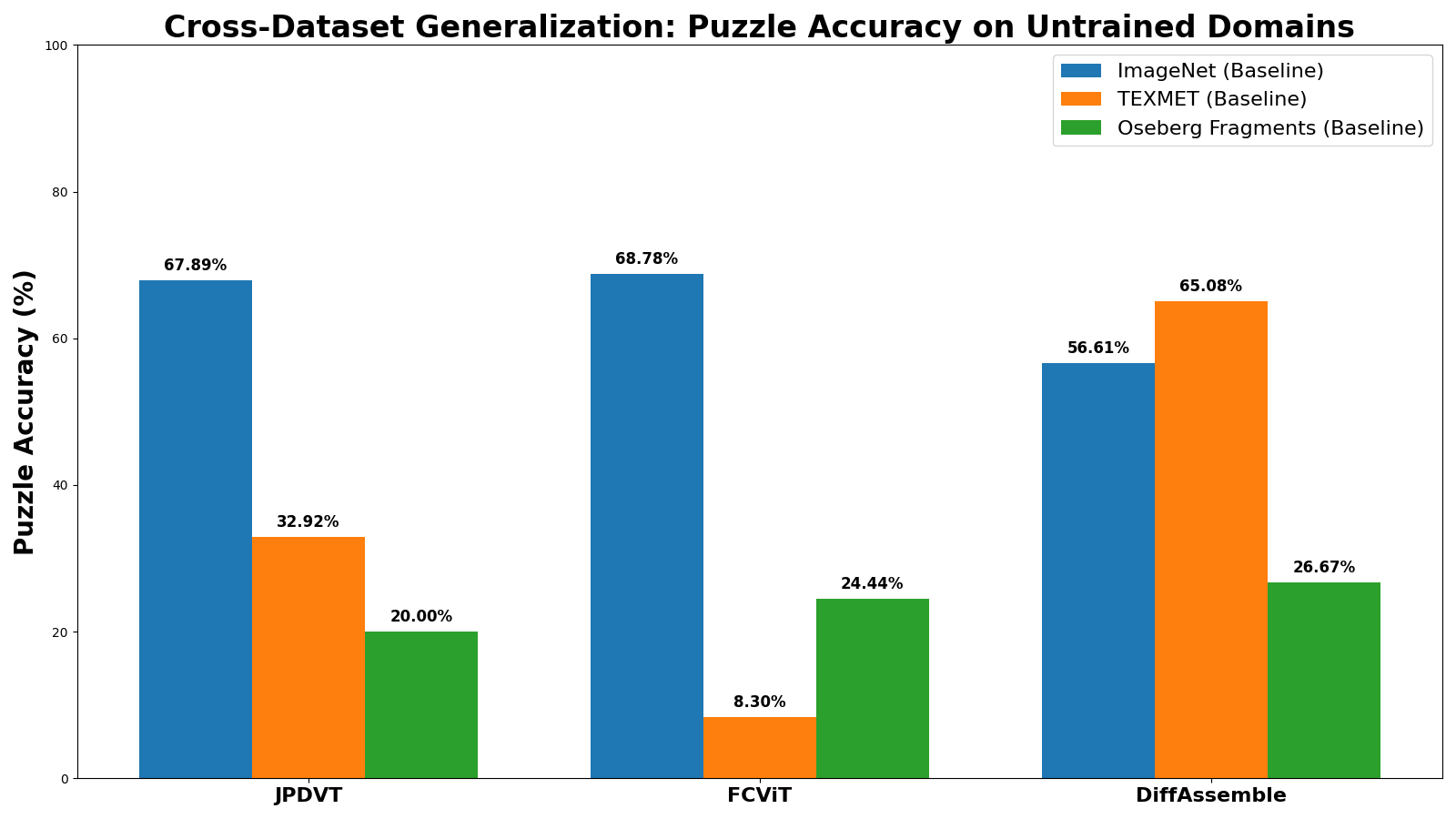}
    \caption{Cross-dataset evaluation comparing model performance on ImageNet, TEXMET, and Oseberg fragments. DiffAssemble maintains performance while FCViT collapses, demonstrating the domain gap between natural images and historical textiles.}
    \label{fig:cross-domain}
    \Description{Grouped bar chart of puzzle accuracy for three models on three datasets. JPDVT scores 67.89 percent on ImageNet, 32.92 percent on TEXMET and 20.00 percent on Oseberg fragments. FCViT scores 68.78, 8.30 and 24.44 percent. DiffAssemble scores 56.61, 65.08 and 26.67 percent. FCViT's TEXMET bar is by far the shortest, while DiffAssemble is the only model whose textile accuracy exceeds its ImageNet accuracy.}
\end{figure}

Figure~\ref{fig:cross-domain} visualizes this domain gap, illustrating DiffAssemble's superior robustness to distribution shift compared to transformer-based approaches.

\subsection{Oseberg Fragment Baseline}

We evaluated all models on the 45 Oseberg textile fragments to establish baseline performance on real archaeological materials (see Figure~\ref{fig:oseberg-examples} for examples of 3$\times$3 fragmented Oseberg puzzles). These fragments present extreme challenges: an average masked area of 75.60\% indicates severe degradation with minimal visual context.

\begin{table}[t]
\centering
\caption{Baseline performance on Oseberg fragments (3$\times$3 puzzles, 45 fragments). All models trained on ImageNet without textile-specific fine-tuning.}
\label{tab:oseberg-baseline}
\begin{tabular}{@{}lc@{}}
\toprule
\textbf{Model} & \textbf{Puzzle Accuracy (\%)} \\
\midrule
JPDVT & 20.00 \\
FCViT & 24.44 \\
DiffAssemble & 26.67 \\
\bottomrule
\end{tabular}
\end{table}

Table~\ref{tab:oseberg-baseline} shows that all models struggle significantly on Oseberg fragments, with accuracies ranging from 20.00\% to 26.67\%. DiffAssemble achieved the highest baseline at 26.67\%, consistent with its superior cross-domain robustness observed on TEXMET. These low accuracies confirm that ImageNet pre-training alone is insufficient for archaeological textile reconstruction.

\subsection{Domain Adaptation via Fine-Tuning}
\label{sec:finetuning}

Based on the cross-domain results, we investigated whether domain-specific fine-tuning could improve performance. DiffAssemble was selected for fine-tuning experiments given its superior baseline performance and architectural flexibility.

\begin{table}[t]
\centering
\caption{DiffAssemble performance before and after fine-tuning on TEXMET. The model was fine-tuned for 150 epochs on TEXMET training data.}
\label{tab:finetuning}
\begin{tabular}{@{}lccc@{}}
\toprule
\textbf{Evaluation Dataset} & \textbf{Baseline (\%)} & \textbf{Fine-tuned (\%)} & \textbf{Improvement} \\
\midrule
TEXMET & 65.08 & 85.89 & +20.81 \\
Oseberg & 26.67 & 64.44 & +37.77 \\
\bottomrule
\end{tabular}
\end{table}

Table~\ref{tab:finetuning} presents the key finding of this work: domain-specific fine-tuning dramatically improves performance on both textile datasets. On TEXMET, accuracy improved from 65.08\% to 85.89\%, a gain of 20.81 percentage points. Most significantly, on the challenging Oseberg fragments, accuracy improved from 26.67\% to 64.44\% (a 37.77 percentage point improvement), demonstrating that models can learn to handle highly degraded archaeological materials when provided with appropriate training data.

\begin{figure}[t]
    \centering
    \includegraphics[width=\columnwidth]{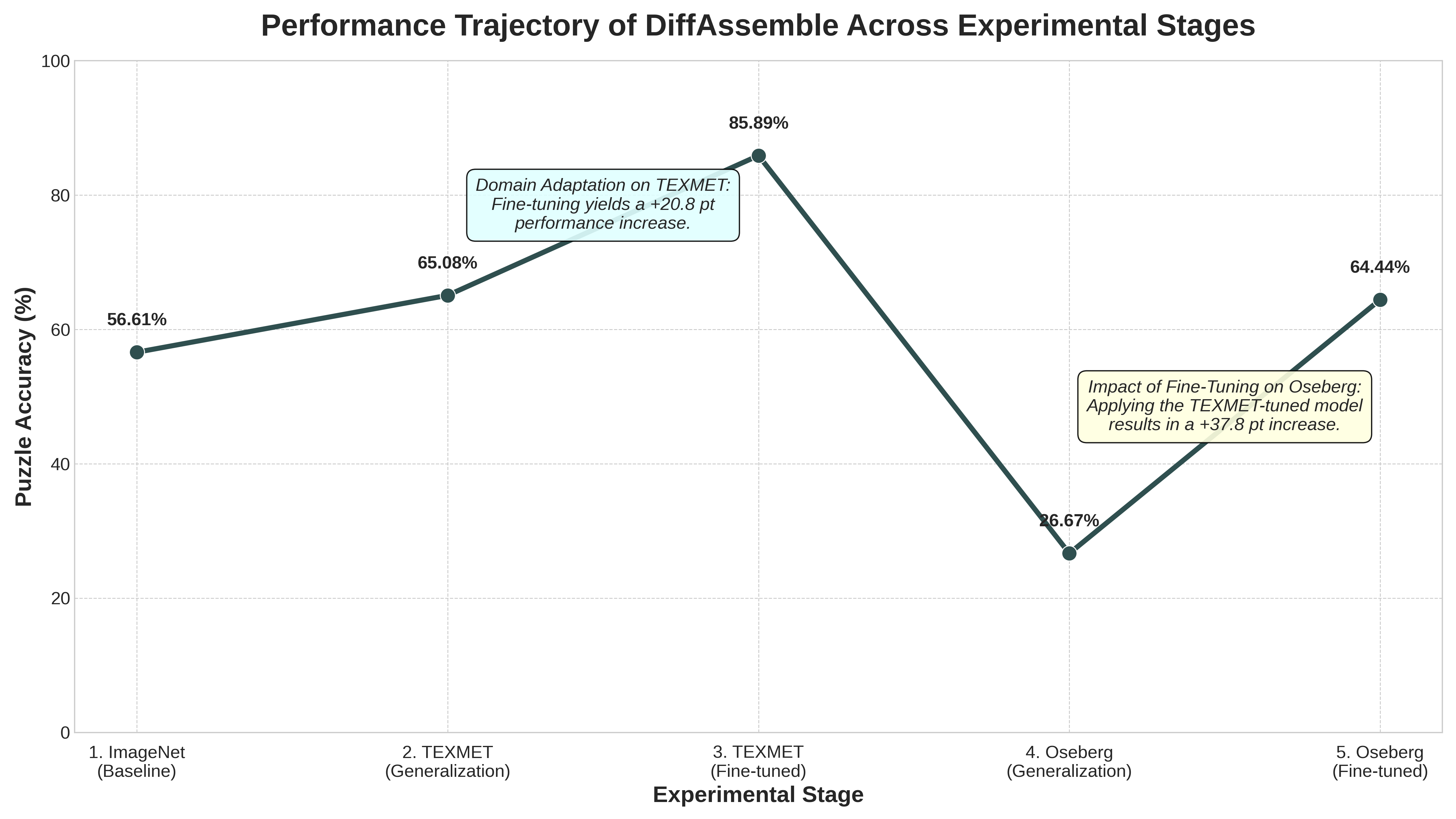}
    \caption{Summary of DiffAssemble puzzle accuracy across the five experimental stages. Fine-tuning on TEXMET lifts accuracy in both evaluation domains: TEXMET rises from 65.08\% to 85.89\% and the held-out Oseberg fragments from 26.67\% to 64.44\%, indicating that the learned representations transfer to degraded archaeological materials. The drop between the third and fourth stages reflects the switch to the much harder Oseberg domain, not a decline in the model.}
    \label{fig:trajectory}
    \Description{Line chart of DiffAssemble puzzle accuracy across five experimental stages: ImageNet baseline 56.61 percent, TEXMET generalization 65.08 percent, TEXMET fine-tuned 85.89 percent, Oseberg generalization 26.67 percent and Oseberg fine-tuned 64.44 percent. The line climbs to a peak at the third stage, drops sharply at the fourth stage where the evaluation domain changes to Oseberg, then rises again. Two callout boxes note a gain of 20.8 points from fine-tuning on TEXMET and a gain of 37.8 points on Oseberg.}
\end{figure}

Figure~\ref{fig:trajectory} summarizes performance across the five experimental stages, showing that fine-tuning improves accuracy on both TEXMET and the held-out Oseberg evaluation set.

Fine-tuning experiments with JPDVT and FCViT were unsuccessful. JPDVT exhibited catastrophic forgetting~\cite{McCloskey1989}, its TEXMET accuracy falling from 32.92\% to 10.88\%, while FCViT improved only marginally (8.30\% to 10.86\%), indicating limited adaptation capacity rather than successful domain transfer. This suggests that DiffAssemble's graph-based architecture may be particularly well-suited to domain transfer.

\subsection{Analysis: What Did the Model Learn?}

A critical question is whether the fine-tuned model learned textile-specific visual patterns or merely exploited shape information from puzzle piece boundaries. To investigate this, we evaluated using inscribed square crops that remove corner shape cues while preserving central textile content.

\begin{table}[t]
\centering
\caption{Inscribed cropping analysis for DiffAssemble: puzzle accuracy (\%) with standard and inscribed crops. Inscribed crops remove shape information from puzzle piece boundaries.}
\label{tab:inscribed}
\begin{tabular}{@{}llcc@{}}
\toprule
\textbf{Dataset} & \textbf{Training} & \textbf{Standard} & \textbf{Inscribed} \\
\midrule
TEXMET & ImageNet baseline & 65.08 & 25.79 \\
TEXMET & Fine-tuned & 85.89 & 27.08 \\
Oseberg & ImageNet baseline & 26.67 & 27.67 \\
Oseberg & Fine-tuned & 64.44 & 39.81 \\
\bottomrule
\end{tabular}
\end{table}

Table~\ref{tab:inscribed} shows that for the fine-tuned model, performance decreases from 64.44\% to 39.81\% on Oseberg when shape information is removed. The effect is even stronger on TEXMET, where the fine-tuned model falls from 85.89\% to 27.08\%, indicating that on intact textile images the model leans more heavily on boundary cues than on the degraded fragments. Conversely, the ImageNet baseline on Oseberg improves slightly under inscribed cropping (26.67\% to 27.67\%), showing that removing ragged fragment boundaries does not simply destroy information. While these drops indicate that shape cues contribute significantly to performance, all inscribed results remain far above random chance (a random arrangement solves a complete 3$\times$3 puzzle with probability $1/9! \approx 3\times10^{-4}$\%, and the corresponding fragment-level random baseline is $1/9 \approx 11.1$\%), demonstrating that the model has learned meaningful textile patterns beyond mere shape matching.

\begin{figure}[t]
    \centering
    \includegraphics[width=\columnwidth]{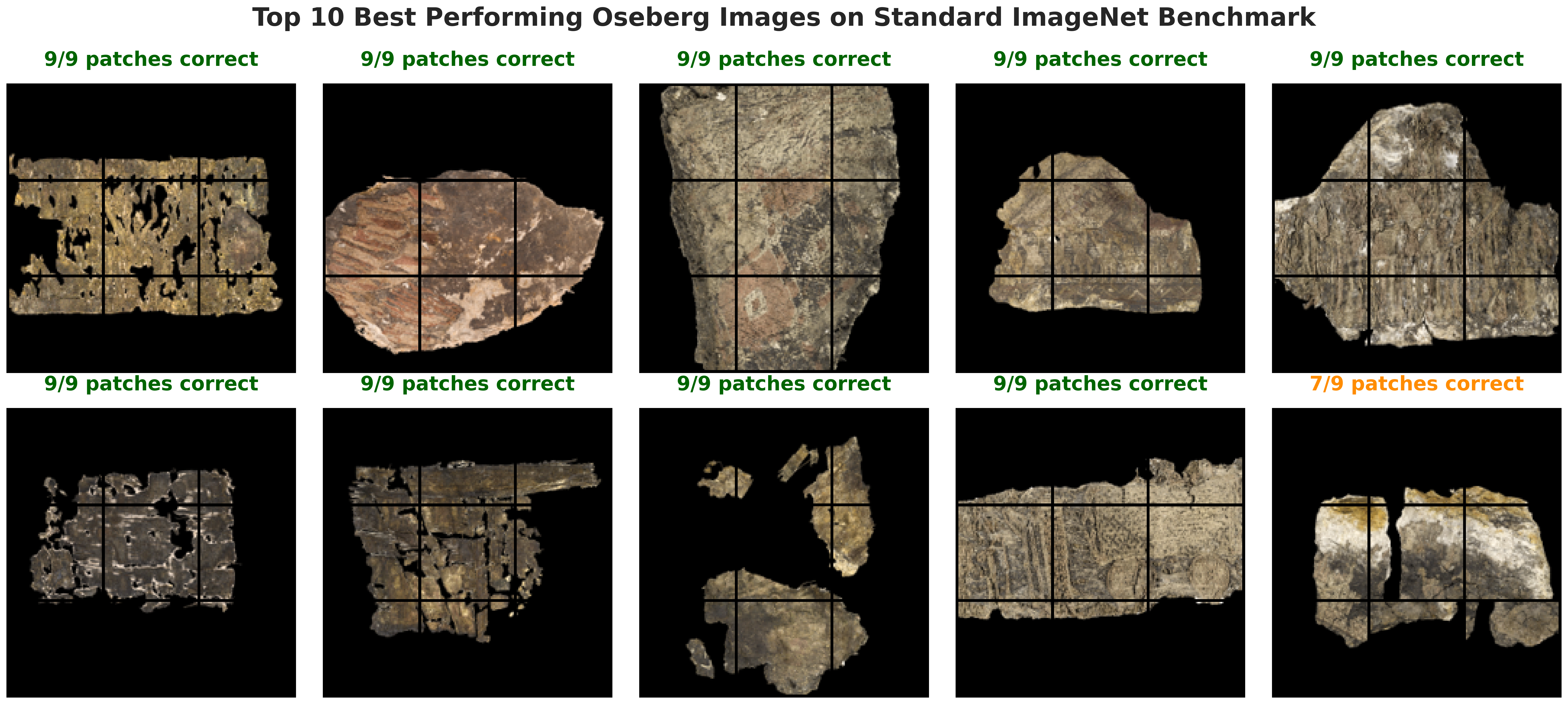}
    \caption{The ten Oseberg fragment images best reordered by an ImageNet-trained baseline model. Each fragment photograph is divided into a 3$\times$3 grid and reassembled in isolation; nine of the ten are restored completely (9/9 patches) and one partially (7/9). Fragments with distinctive global structure survive severe degradation even under the baseline model, within this simplified setting.}
    \label{fig:oseberg-examples}
    \Description{Grid of ten photographs of Oseberg textile fragments on black backgrounds, each overlaid with a 3 by 3 grid marking how it was split and reassembled. Nine are labeled 9 of 9 patches correct in green and the tenth is labeled 7 of 9 patches correct in orange. The fragments are irregular in outline and heavily degraded, with woven texture and surface loss clearly visible.}
\end{figure}

Figure~\ref{fig:oseberg-examples} presents the Oseberg fragment images most successfully reordered by the baseline model, illustrating that fragments with distinctive global structure remain solvable despite severe degradation.

\subsection{Impact of Masking and Inpainting}
\label{sec:masking}

To investigate how degradation patterns affect puzzle-solving performance, we conducted experiments with different masking strategies, on a curated subset of 1,000 images from the ImageNet validation set providing a controlled visual foundation, and evaluated whether inpainting could recover lost information.

\textbf{Masking Strategies.} We evaluated two masking approaches, both calibrated to obscure approximately 35.6\% of the image area: (1) regular centered rectangular masking simulating uniform degradation, and (2) irregular masking derived from a single representative Oseberg fragment boundary, resized and applied to every image so that all images share the same degradation pattern, representing realistic archaeological damage patterns.

Table~\ref{tab:inpainting-results} summarizes the impact of masking and inpainting on puzzle accuracy across all three models.

\begin{table}[t]
\centering
\caption{Impact of masking and inpainting on 3$\times$3 puzzle accuracy (\%), evaluated on 1,000 ImageNet validation images per condition.}
\label{tab:inpainting-results}
\begin{tabular}{@{}lccc@{}}
\toprule
\textbf{Condition} & \textbf{JPDVT} & \textbf{FCViT} & \textbf{DiffAssemble} \\
\midrule
No mask (baseline) & 87.80 & 70.62 & 55.70 \\
Regular mask & 87.30 & 60.83 & 50.50 \\
Irregular mask & 79.10 & 46.15 & 54.70 \\
Regular + inpainting & 86.10 & 65.10 & 58.50 \\
Irregular + inpainting & 84.30 & 60.10 & 57.90 \\
\bottomrule
\end{tabular}
\end{table}

 One caveat on reading the table: on the 1,000-image subset, JPDVT's no-mask baseline (87.80\%) sits well above its full-split baseline of 67.89\% (Section~\ref{sec:imagenet-baseline}), while the other two models land within two points of theirs; masking effects should therefore be read against the table's own baseline row, not across sections. The three architectures respond to degradation in very different ways. JPDVT is essentially unaffected by regular masking (87.30\% against its 87.80\% baseline): the regular grid mask, aligned with the image coordinates, removes information without disturbing the patch structure the model relies on. Irregular, archaeologically derived masks are the harder condition for JPDVT, costing 8.7 percentage points. FCViT is the most vulnerable model, losing 24.5 points under irregular masking, consistent with its reliance on global visual features that scattered occlusion destroys. DiffAssemble starts from a lower baseline but is the most robust, varying within a band of roughly 8 points across all conditions; its graph-based, relational formulation appears inherently less sensitive to missing content. We note that repeated JPDVT evaluations with different random seeds varied by up to 6.4 percentage points under regular masking; we report the values consistent with the underlying experiment logs.

\begin{figure}[t]
    \centering
    \includegraphics[width=\columnwidth]{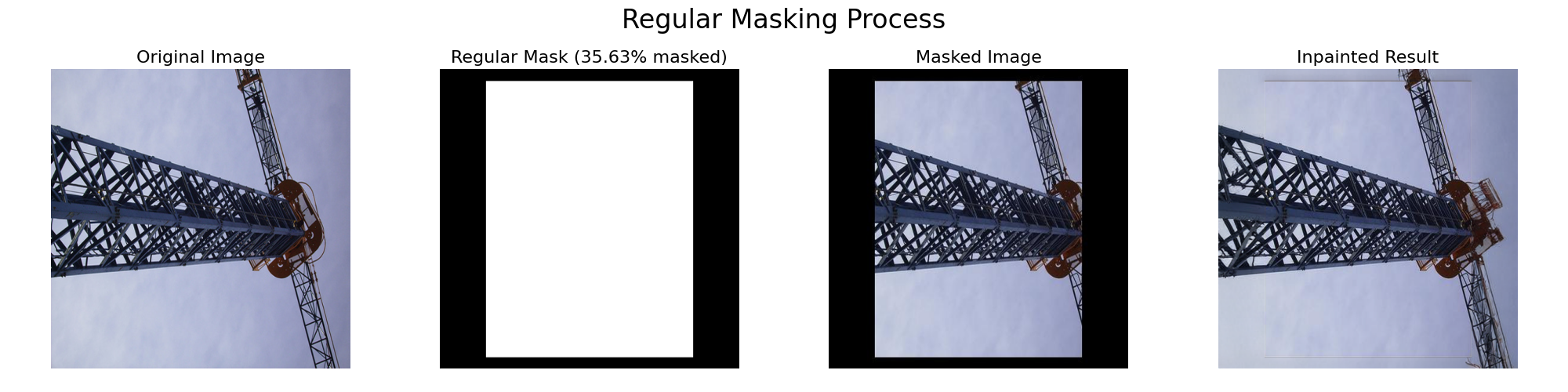}\\
    \includegraphics[width=\columnwidth]{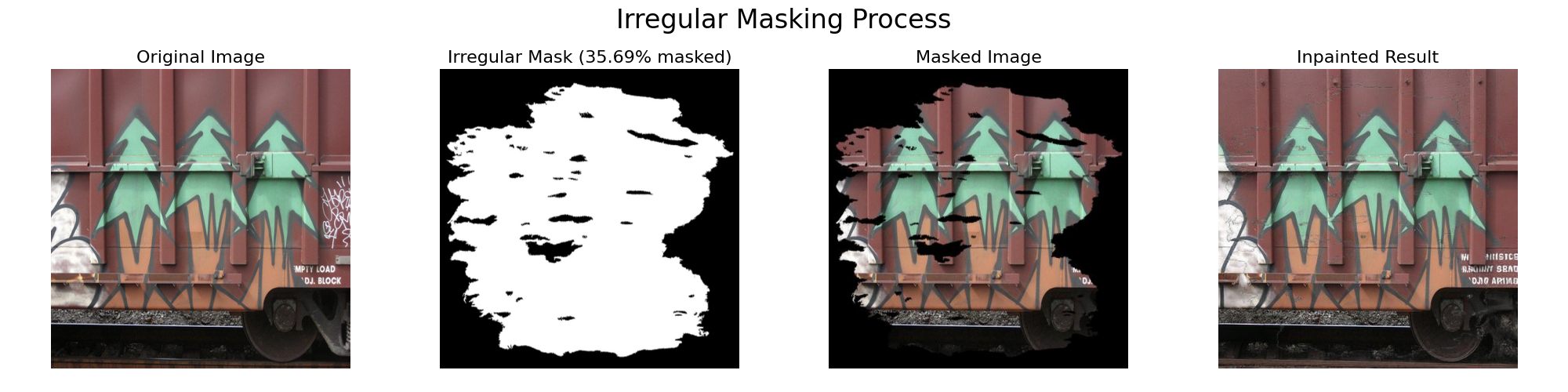}
    \caption{The masking and inpainting pipeline on ImageNet validation images. Top: regular centered-rectangle mask. Bottom: irregular mask derived from a single Oseberg fragment. Panels left to right: original image, mask, masked image, and ASUKA-inpainted result. The inpainted images restore visual plausibility to the masked regions.}
    \label{fig:inpainting-example}
    \Description{Two rows of four panels showing the masking and inpainting pipeline on ImageNet validation images. The top row uses a photograph of a construction crane with a regular centered rectangular mask covering 35.63 percent of the image. The bottom row uses a photograph of a graffiti-covered freight wagon with an irregular blob-shaped mask derived from an Oseberg fragment, covering 35.69 percent. In each row the panels are, from left to right, the original image, the binary mask, the masked image with the removed region shown in black, and the ASUKA-inpainted result, which restores a visually plausible scene.}
\end{figure}

\textbf{Inpainting Recovery.} We applied ASUKA~\cite{ASUKA2025}, a state-of-the-art inpainting model, to masked images before puzzle solving. Figure~\ref{fig:inpainting-example} illustrates the full pipeline for both mask types, from the original image to the inpainted result; quantitative fidelity metrics, which show that irregular masks are consistently harder to inpaint, are reported in the supplementary material.
Figure~\ref{fig:inpainting-effect} shows that inpainting is a useful but not universal recovery strategy, and its benefit depends on both the model and the mask type. It helps most where the damage hurts most: FCViT recovers 14.0 points on irregularly masked images and JPDVT recovers 5.2, while DiffAssemble gains a more modest 3.2. On regularly masked images the picture is mixed: JPDVT's small change (86.10\% against 87.30\%) lies within the seed variability documented in the supplementary material (up to 6.4 points), so we do not interpret it, while FCViT and DiffAssemble still benefit. Inpainting cannot fully restore baseline performance for any model under realistic irregular degradation.

\begin{figure}[t]
    \centering
    \includegraphics[width=\columnwidth]{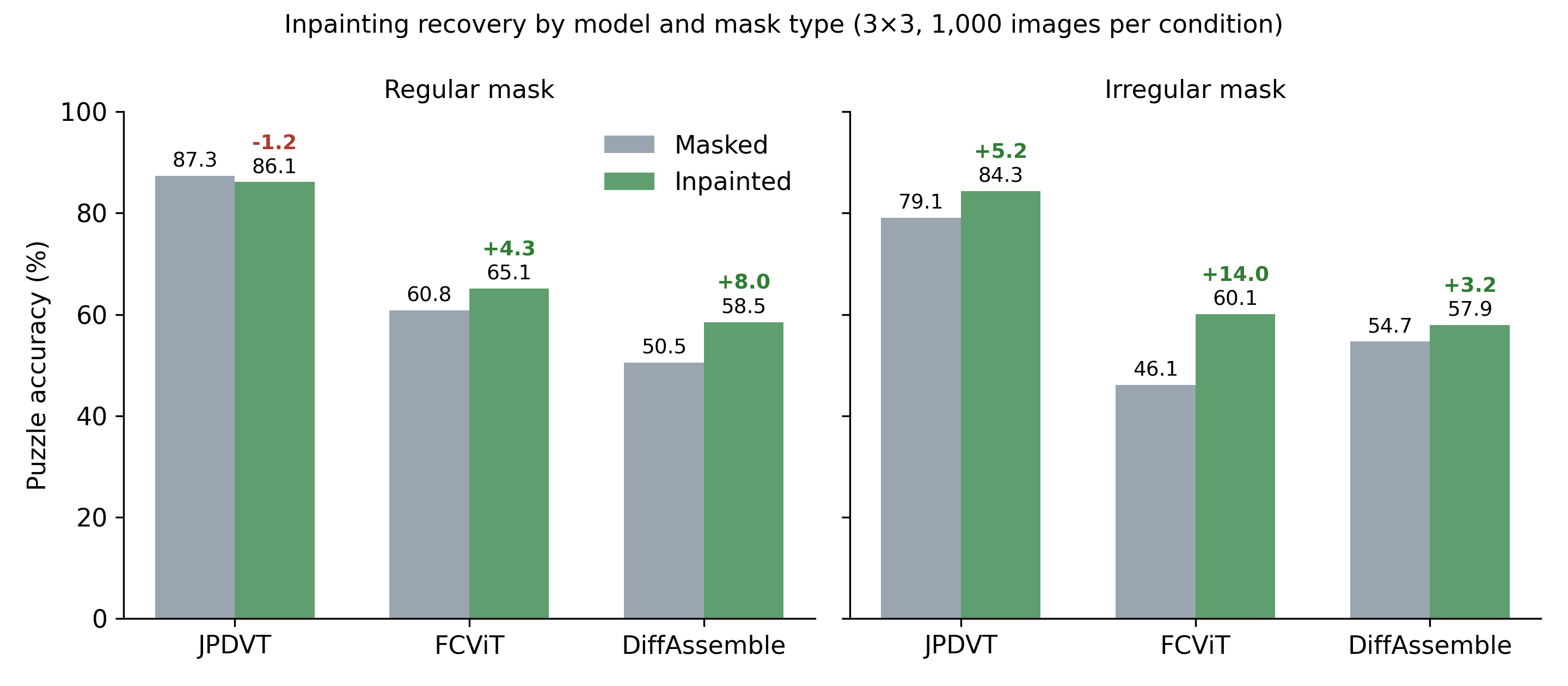}
    \caption{Inpainting effectiveness: comparison of puzzle accuracy before (masked) and after inpainting for each model under both masking conditions.}
    \label{fig:inpainting-effect}
    \Description{Two grouped bar charts comparing puzzle accuracy on masked images against inpainted images for each model, using 1,000 images per condition at 3 by 3. Under the regular mask, JPDVT moves from 87.3 to 86.1 percent (down 1.2 points), FCViT from 60.8 to 65.1 percent (up 4.3) and DiffAssemble from 50.5 to 58.5 percent (up 8.0). Under the irregular mask, JPDVT moves from 79.1 to 84.3 percent (up 5.2), FCViT from 46.1 to 60.1 percent (up 14.0) and DiffAssemble from 54.7 to 57.9 percent (up 3.2).}
\end{figure}

These results suggest that while inpainting can partially reduce realistic degradation effects, most strongly for architectures that depend on global visual features, the domain-specific fine-tuning approach (Section~\ref{sec:finetuning}) provides more substantial improvements for archaeological materials.

\subsection{Ablation Analysis}
\label{sec:ablation}

To isolate the factors that drive reassembly performance on TEXMET, we analyze three variables: fragment count (grid size), material category as a proxy for visual complexity, and degradation type (Section~\ref{sec:masking}).

\textbf{Fragment count.}
Table~\ref{tab:ablation-grid} reports JPDVT puzzle accuracy as the grid grows from $3\times3$ (9 fragments) to $4\times4$ (16 fragments). The same transition costs JPDVT four times more on textiles than on natural images: as tiles shrink, repetitive weave patterns provide even fewer distinctive cues per tile, so fragment-count sensitivity is strongly amplified by domain difficulty. The other two architectures diverge from this finding rather than repeat it: DiffAssemble's checkpoints differ in training budget across grid sizes, so no controlled cross-grid comparison is possible for it; FCViT cannot be evaluated beyond $3\times3$ at all, since its coordinate-regression head is trained for a fixed nine-fragment layout. The grid constraint itself is thus not only a benchmark simplification but a live architectural boundary of the model class.

\begin{table}[t]
\centering
\caption{Grid-size ablation: JPDVT puzzle accuracy (\%) as fragment count increases from $3\times3$ (9 fragments) to $4\times4$ (16 fragments). The $4\times4$ configurations use the JPDVT-T variant (patch size 64, 256$\times$256 input) under the same evaluation protocol; TEXMET values are computed on the 1,865-image test split.}
\label{tab:ablation-grid}
\begin{tabular}{@{}lccc@{}}
\toprule
\textbf{Dataset} & \textbf{$3\times3$} & \textbf{$4\times4$} & \textbf{Relative drop} \\
\midrule
ImageNet & 67.89 & 56.62 & $-17\%$ \\
TEXMET   & 32.92 & 9.54  & $-71\%$ \\
\bottomrule
\end{tabular}
\end{table}

\textbf{Material category.}
Because expert weave-type labeling of 18,644 objects is impractical, we group the test split by the museum's curatorial classification as a proxy for visual complexity. Table~\ref{tab:ablation-material} shows a $2.9\times$ spread in JPDVT baseline accuracy between the easiest and hardest categories: pictorial tapestries (53.3\%) provide globally distinctive scene content, whereas printed textiles (18.1\%) are dominated by flat repeating patterns whose tiles are locally interchangeable. Woven textiles, the largest group ($n{=}883$), fall below the overall mean (29.4\% against 32.9\%), consistent with the qualitative failure analysis: repetitive weave structure leaves the solver without positional landmarks. The same ordering holds at $4\times4$ (tapestries 18.3\%, printed 4.4\%), indicating that the category effect is a property of the material rather than of a particular grid size.

\begin{table}[t]
\centering
\caption{Material-category ablation: ImageNet-trained JPDVT evaluated on TEXMET $3\times3$ (the cross-domain setting of Table~\ref{tab:cross-domain}), named material categories with $n\geq20$ test images, MET curatorial classification used as a complexity proxy. The residual Other group is excluded here and reported in the supplementary material.}
\label{tab:ablation-material}
\begin{tabular}{@{}lccc@{}}
\toprule
\textbf{Category} & $n$ & \textbf{Puzzle Acc.\ (\%)} & \textbf{Fragment Acc.\ (\%)} \\
\midrule
Tapestries  & 60  & 53.3 & 71.9 \\
Embroidered & 350 & 37.7 & 58.9 \\
Costume     & 34  & 35.3 & 62.1 \\
Rugs        & 36  & 33.3 & 56.8 \\
Woven       & 883 & 29.4 & 53.0 \\
Velvets     & 121 & 28.9 & 60.0 \\
Printed     & 160 & 18.1 & 44.0 \\
\bottomrule
\end{tabular}
\end{table}

\textbf{Degradation type.}
The masking and inpainting analysis (Section~\ref{sec:masking}, Table~\ref{tab:inpainting-results}) completes the ablation: irregular, archaeologically derived masks harm JPDVT and FCViT substantially more than regular masks, the graph-based DiffAssemble is the least sensitive to either, and the benefit of inpainting is largest exactly where degradation hurts most.

\section{Discussion}
\label{sec:discussion}

\subsection{Interpretation of Results}

The substantial performance differences between architectures provide insight into what makes a puzzle-solving model effective for textile puzzle reassembly.

\textbf{Why DiffAssemble Succeeded.} DiffAssemble's graph-based representation proved particularly well-suited for textile data. By encoding fragments as nodes and spatial relationships as edges, the model can capture the local connectivity patterns that characterize woven structures. The diffusion-based refinement process allows iterative improvement of predictions, which appears beneficial for degraded materials where initial feature extraction may be uncertain. Additionally, DiffAssemble's ability to handle rotation and missing pieces aligns well with the realities of archaeological materials.

\begin{figure}[t]
    \centering
    \includegraphics[width=\columnwidth]{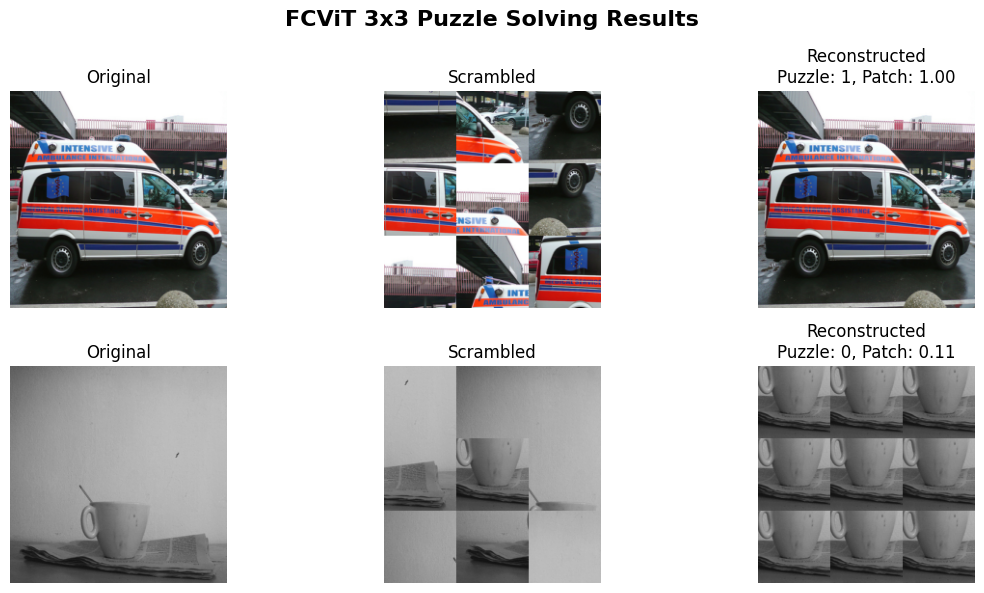}
    \caption{FCViT 3$\times$3 puzzle-solving results on ImageNet~\cite{ImageNet2009}. Top: successful reconstruction (9/9 pieces correct). Bottom: failure case showing the duplicate placement problem, where multiple fragments are assigned to the same coordinates (1/9 pieces correct).}
    \label{fig:fcvit-duplicates}
    \Description{Two rows of three panels showing FCViT results on ImageNet 3 by 3 puzzles, with the original image, the scrambled version and the reconstruction in each row. In the top row a color photograph of an ambulance is reassembled correctly, with patch accuracy 1.00. In the bottom row a grayscale photograph of a cup on a newspaper is reconstructed with the same fragment repeated across the grid, giving patch accuracy 0.11, which illustrates the duplicate placement failure.}
\end{figure}

\textbf{Why FCViT Struggled.} FCViT's coordinate regression approach showed severe sensitivity to domain shift, collapsing from competitive ImageNet performance to 8.30\% on TEXMET. The regression formulation lacks explicit uniqueness constraints, making it prone to duplicate placements where multiple fragments are assigned similar coordinates (Figure~\ref{fig:fcvit-duplicates}). This limitation becomes critical when visual features are degraded or unfamiliar, as the model lacks the structured constraints that could guide predictions in the absence of strong visual signals.

\textbf{Domain Adaptation Finding.} The most significant finding is that domain-specific fine-tuning can overcome much of the domain gap. The 37.77 percentage point improvement on Oseberg fragments (from 26.67\% to 64.44\%) demonstrates that with appropriate training data, models can learn to handle the unique challenges of archaeological textiles: faded colors, degraded fibers, and missing content. This suggests that the fundamental limitations were not architectural but rather stemmed from training data that did not represent the target domain.

\subsection{Practical Implications}

For practitioners, three implications follow. TEXMET provides the first large-scale, CC0-licensed dataset for textile puzzle-solving research, enabling unrestricted academic and commercial use in heritage preservation. Among the evaluated architectures, our results favor DiffAssemble's graph-based approach, whose robustness to domain shift and ease of fine-tuning make it the most suitable current choice for textile imagery. Finally, domain-specific training proves essential: models trained solely on natural images cannot reliably transfer to textile materials, and even limited domain-specific data substantially improves performance on imagery of real archaeological artifacts.

\subsection{Fragment Realism and Degradation Effects}
\label{sec:realism}

A fair question about any benchmark built from synthetic puzzles is how its difficulty relates to the condition of real archaeological material. No established quantitative measure exists for the realism of a synthetically generated fragment: realism spans several dimensions at once, including shape irregularity, missing area, edge damage, fading, staining, and weave distortion. We therefore assess realism qualitatively, and anchor the evaluation with real material. The contrast can be stated plainly: TEXMET tiles have zero masked area and identical, regular square geometry by construction, whereas the real Oseberg fragments average 75.60\% masked area with a mean compactness of 0.15 (isoperimetric compactness, $4\pi A/P^2$, where $A$ is the visible fragment area and $P$ its perimeter; 1.0 for a perfect disc, lower for irregular outlines).

The regular-grid formulation itself is a constraint of the current model class, not a choice of this benchmark. JPDVT and FCViT accept only regular square grids with a fixed piece count, and FCViT additionally supports no rotation. DiffAssemble supports rotations and missing pieces, and its graph formulation could in principle extend to irregular outlines, but no trained model for irregular fragment geometry exists. We additionally attempted a more realistic distractor experiment, presenting 16 candidate fragments of which only 9 belong to the puzzle; this failed at the implementation stage for all three models, whose architectures assume a fixed input fragment count. Grid puzzles are, at present, the evaluation this model class supports, which is why we position the benchmark as a controlled first step.

What the evaluation does include is real degradation. The 45 Oseberg fragment images carry genuine archaeological damage: the fading, staining, holes, and irregular outlines visible in Figure~\ref{fig:oseberg_fragments}. Reassembly performance on fragments with larger missing areas is systematically worse, directly linking degradation severity to task difficulty. Most importantly, the fine-tuned model transfers from synthetic training puzzles to this genuinely degraded imagery with a substantial gain (26.67\% to 64.44\%), evidence that the benchmark's training signal is relevant beyond its synthetic construction.

Our qualitative assessment is consistent with independent evidence from a parallel line of work. Berels~\cite{berels2025deep} investigated deep feature extraction for content-based retrieval of archaeological textile fragments under irregular masking conditions, explicitly targeting the gap between theoretical puzzle-solving approaches and the practical condition of textile artifacts, which, unlike rigid materials such as pottery, decompose irregularly into fragments with ragged threaded edges and holes. Her evaluation of CNN and ViT feature extractors shows that retrieval performance decreases as masking grows more complex and irregular: with increasing degradation the extracted features become less informative about the depicted content, as the models increasingly respond to the degradation pattern itself rather than to the underlying textile. Moreover, Berels argues that a smaller part of the fragment cropped to a regular size may be more informative than the image of an entire fragment which also depicts degraded areas, since the latter act as a distractor and introduce noise. She concludes that fully automated puzzle solving is not yet feasible for heavily degraded archaeological artifacts whose fragment boundaries no longer match, and instead recommends retrieval-based assistance to archaeologists. These findings mirror our observations at the assembly level: performance degrades most under irregular, archaeologically derived masking (Section~\ref{sec:masking}), and inpainting recovers only part of the loss. Together, the two studies suggest that sensitivity to realistic degradation is a property of current deep visual representations in general, rather than an artifact of any single task formulation.

The degradation analysis in Section~\ref{sec:masking} quantifies these effects by type: irregular, archaeologically derived masks are consistently harder than regular grid masks for the transformer-based models, and inpainting recovers only part of the loss, helping most where degradation hurts most. The gap between regular and irregular masking conditions is, in one number, the gap between synthetic and archaeological difficulty. Closing it will require architectures that accept irregular fragment geometry, reason across incomplete fragment sets, and match fragments across, rather than within, textiles.

\subsection{Limitations}

Several limitations constrain the interpretation and applicability of our findings:

\textbf{Grid-Based Puzzle Formulation.} Our evaluation uses regular 3$\times$3 grid puzzles, whereas real archaeological fragments have irregular shapes resulting from damage and degradation. Extending these methods to irregular fragment geometries remains an open challenge.

\textbf{Simplified Oseberg Evaluation.} No ground truth exists for how the Oseberg fragments relate to each other in the original tapestry, making cross-fragment matching evaluation impossible. Instead, we artificially split each surviving fragment into a 3$\times$3 grid of smaller patches to create known ground truth, enabling reliable evaluation of puzzle-solving performance. Consequently, we evaluated individual fragments in isolation rather than attempting full tapestry reconstruction. The real archaeological challenge involves simultaneously reasoning about hundreds of fragments that may belong to multiple distinct textiles.

\textbf{Limited Archaeological Scale.} Our evaluation used 45 Oseberg fragments. While this represents a significant step toward real-world application, larger-scale evaluation would strengthen the findings.

\textbf{Synthetic Fragmentation.} TEXMET uses simulated puzzle cuts on intact textile images. Real archaeological fragmentation involves irregular damage, fiber loss, and material degradation that synthetic puzzles cannot fully capture. The gap between simulated and real fragmentation may limit transfer to the most challenging archaeological cases.

\section{Conclusion}
\label{sec:conclusion}

This paper addressed the challenge of applying deep learning puzzle-solving methods to archaeological textile imagery, as a step toward computational reconstruction, using the Viking Age Oseberg textile fragments as a case study.

\subsection{Contributions}

We presented four main contributions:

\begin{enumerate}
    \item \textbf{TEXMET Dataset:} We introduced TEXMET, a dataset of 18,644 curated historical textile images from the Metropolitan Museum of Art's Open Access collection. To our knowledge, this represents the first textile-specific benchmark for puzzle-solving research.

    \item \textbf{Systematic Evaluation:} We provided the first systematic evaluation of three state-of-the-art puzzle-solving architectures (JPDVT, FCViT, and DiffAssemble) on both a large textile dataset and real archaeological fragments from the Oseberg ship burial.

    \item \textbf{Domain Adaptation Study:} We demonstrated that fine-tuning on domain-specific data is essential for archaeological applications. DiffAssemble improved from 26.67\% to 64.44\% puzzle accuracy on Oseberg fragments after fine-tuning on TEXMET, a 37.77 percentage point improvement.

    \item \textbf{Open-Source Release:} All code, experimental frameworks, trained model checkpoints, and the TEXMET dataset will be made publicly available to support reproducibility and future research.
\end{enumerate}

\subsection{Key Findings}

Our experiments revealed several important findings for the application of computational methods to cultural heritage reconstruction:

\textbf{Domain gap is substantial.} Transformer-based models trained on natural images fail to generalize to textile data. FCViT collapsed from competitive ImageNet performance to 8.30\% on TEXMET, and JPDVT fell from 67.89\% to 32.92\%; only the graph-based DiffAssemble carried its performance level across the shift (65.08\% on TEXMET against a 56.61\% ImageNet baseline).

\textbf{Architecture choice matters.} DiffAssemble's graph-based approach demonstrated superior robustness to domain shift compared to pure transformer architectures. Its ability to handle rotation, missing pieces, and variable puzzle sizes makes it in principle particularly suitable for archaeological applications, although the present benchmark exercises only fixed grids without rotation.

\textbf{Domain-specific training is essential.} The dramatic improvement from fine-tuning (particularly the 37.77 percentage point gain on highly degraded Oseberg fragments) demonstrates that appropriate training data can enable effective application of deep learning to archaeological materials.

\textbf{Models learn meaningful patterns.} The inscribed cropping analysis confirmed that fine-tuned models learn textile-specific visual features rather than merely exploiting shape information, suggesting genuine pattern recognition capability.

\subsection{Future Work}

Several directions remain for future research:

\textbf{Irregular Fragment Handling.} Extending puzzle-solving methods beyond regular grid configurations to handle the irregular shapes of real archaeological fragments represents a critical next step toward practical application.

\textbf{Full Tapestry Reconstruction.} Scaling from individual fragment puzzles to simultaneous reconstruction of multiple textiles from mixed fragment collections would address the realistic archaeological scenario.

\textbf{Broader Archaeological Applications.} The domain adaptation approach demonstrated here may generalize to other fragmented heritage materials including ceramics, frescoes, and manuscripts.

\textbf{Expanded Datasets.} Creating larger and more diverse textile datasets, potentially incorporating real archaeological fragment imagery, would further improve model robustness and enable evaluation at archaeological scale.

The combination of TEXMET for training and DiffAssemble's graph-based architecture provides a foundation for computational approaches to textile heritage preservation. As these methods mature, they offer the potential to accelerate the careful, time-consuming work of archaeological reconstruction while providing new insights into historical material culture.

\subsection{Data Availability}

All source code, experimental frameworks, trained model checkpoints, and the TEXMET dataset will be made publicly available.

\begin{acks}
The Oseberg textile fragment images were captured by George Alexis Pantos at the Museum of Cultural History, University of Oslo. We thank the Metropolitan Museum of Art for making their Open Access collection available under CC0 licensing. Model training and evaluation were performed on the Idun high-performance computing cluster at NTNU.
\end{acks}

\bibliographystyle{ACM-Reference-Format}
\bibliography{references}


\begin{thebibliography}{41}


\ifx \showCODEN    \undefined \def \showCODEN     #1{\unskip}     \fi
\ifx \showISBNx    \undefined \def \showISBNx     #1{\unskip}     \fi
\ifx \showISBNxiii \undefined \def \showISBNxiii  #1{\unskip}     \fi
\ifx \showISSN     \undefined \def \showISSN      #1{\unskip}     \fi
\ifx \showLCCN     \undefined \def \showLCCN      #1{\unskip}     \fi
\ifx \shownote     \undefined \def \shownote      #1{#1}          \fi
\ifx \showarticletitle \undefined \def \showarticletitle #1{#1}   \fi
\ifx \showURL      \undefined \def \showURL       {\relax}        \fi
\providecommand\bibfield[2]{#2}
\providecommand\bibinfo[2]{#2}
\providecommand\natexlab[1]{#1}
\providecommand\showeprint[2][]{arXiv:#2}

\bibitem[Andersson~Strand(2020)]%
        {Strand2020}
\bibfield{author}{\bibinfo{person}{Eva Andersson~Strand}.}
  \bibinfo{year}{2020}\natexlab{}.
\newblock \showarticletitle{Travelling with Textiles: Production, Consumption,
  and Trade in the {V}iking {A}ge}.
\newblock \bibinfo{journal}{\emph{Bulletin of the Museum of Far Eastern
  Antiquities}}  \bibinfo{volume}{81} (\bibinfo{year}{2020}),
  \bibinfo{pages}{65--88}.
\newblock


\bibitem[Berels(2025)]%
        {berels2025deep}
\bibfield{author}{\bibinfo{person}{Marlena Berels}.}
  \bibinfo{year}{2025}\natexlab{}.
\newblock \emph{\bibinfo{title}{Deep Feature Extraction under Irregular Masking
  Conditions: A Step toward Content-Based Image Retrieval for Archaeological
  Puzzle Solving}}.
\newblock {Master's thesis}. \bibinfo{school}{Norwegian University of Science
  and Technology}.
\newblock


\bibitem[Chen et~al\mbox{.}(2023)]%
        {Chen2023}
\bibfield{author}{\bibinfo{person}{Yingyi Chen}, \bibinfo{person}{Xi Shen},
  \bibinfo{person}{Yahui Liu}, \bibinfo{person}{Qinghua Tao}, {and}
  \bibinfo{person}{Johan A.~K. Suykens}.} \bibinfo{year}{2023}\natexlab{}.
\newblock \showarticletitle{Jigsaw-ViT: Learning Jigsaw Puzzles in Vision
  Transformer}.
\newblock \bibinfo{journal}{\emph{Pattern Recognition Letters}}
  \bibinfo{volume}{166} (\bibinfo{year}{2023}), \bibinfo{pages}{53--60}.
\newblock
\href{https://doi.org/10.1016/j.patrec.2022.12.023}{doi:\nolinkurl{10.1016/j.patrec.2022.12.023}}


\bibitem[Cho et~al\mbox{.}(2010)]%
        {Cho_CVPR_2010}
\bibfield{author}{\bibinfo{person}{Taeg~Sang Cho}, \bibinfo{person}{Shai
  Avidan}, {and} \bibinfo{person}{William~T. Freeman}.}
  \bibinfo{year}{2010}\natexlab{}.
\newblock \showarticletitle{A Probabilistic Image Jigsaw Puzzle Solver}. In
  \bibinfo{booktitle}{\emph{Proc.\ IEEE Conf.\ on Computer Vision and Pattern
  Recognition (CVPR)}}. \bibinfo{pages}{183--190}.
\newblock
\href{https://doi.org/10.1109/CVPR.2010.5540212}{doi:\nolinkurl{10.1109/CVPR.2010.5540212}}


\bibitem[Deng et~al\mbox{.}(2009)]%
        {ImageNet2009}
\bibfield{author}{\bibinfo{person}{Jia Deng}, \bibinfo{person}{Wei Dong},
  \bibinfo{person}{Richard Socher}, \bibinfo{person}{Li-Jia Li},
  \bibinfo{person}{Kai Li}, {and} \bibinfo{person}{Li Fei-Fei}.}
  \bibinfo{year}{2009}\natexlab{}.
\newblock \showarticletitle{ImageNet: A Large-Scale Hierarchical Image
  Database}. In \bibinfo{booktitle}{\emph{Proceedings of the IEEE Conference on
  Computer Vision and Pattern Recognition}}. \bibinfo{pages}{248--255}.
\newblock


\bibitem[Derech et~al\mbox{.}(2021)]%
        {Derech2021}
\bibfield{author}{\bibinfo{person}{Niv Derech}, \bibinfo{person}{Ayellet Tal},
  {and} \bibinfo{person}{Ilan Shimshoni}.} \bibinfo{year}{2021}\natexlab{}.
\newblock \showarticletitle{Solving Archaeological Puzzles}.
\newblock \bibinfo{journal}{\emph{Pattern Recognition}}  \bibinfo{volume}{119}
  (\bibinfo{year}{2021}), \bibinfo{pages}{108065}.
\newblock
\href{https://doi.org/10.1016/j.patcog.2021.108065}{doi:\nolinkurl{10.1016/j.patcog.2021.108065}}


\bibitem[Doersch et~al\mbox{.}(2015)]%
        {Doersch2015}
\bibfield{author}{\bibinfo{person}{Carl Doersch}, \bibinfo{person}{Abhinav
  Gupta}, {and} \bibinfo{person}{Alexei~A Efros}.}
  \bibinfo{year}{2015}\natexlab{}.
\newblock \showarticletitle{Unsupervised visual representation learning by
  context prediction}.
\newblock \bibinfo{journal}{\emph{Proceedings of ICCV}} (\bibinfo{year}{2015}),
  \bibinfo{pages}{1422--1430}.
\newblock


\bibitem[Dosovitskiy et~al\mbox{.}(2021)]%
        {ViT2021}
\bibfield{author}{\bibinfo{person}{Alexey Dosovitskiy}, \bibinfo{person}{Lucas
  Beyer}, \bibinfo{person}{Alexander Kolesnikov}, \bibinfo{person}{Dirk
  Weissenborn}, \bibinfo{person}{Xiaohua Zhai}, \bibinfo{person}{Thomas
  Unterthiner}, \bibinfo{person}{Mostafa Dehghani}, \bibinfo{person}{Matthias
  Minderer}, \bibinfo{person}{Georg Heigold}, \bibinfo{person}{Sylvain Gelly},
  \bibinfo{person}{Jakob Uszkoreit}, {and} \bibinfo{person}{Neil Houlsby}.}
  \bibinfo{year}{2021}\natexlab{}.
\newblock \showarticletitle{An Image is Worth 16x16 Words: Transformers for
  Image Recognition at Scale}. In \bibinfo{booktitle}{\emph{International
  Conference on Learning Representations (ICLR)}}.
\newblock


\bibitem[Freeman and Garder(1964)]%
        {Freeman1964}
\bibfield{author}{\bibinfo{person}{H. Freeman} {and} \bibinfo{person}{L.
  Garder}.} \bibinfo{year}{1964}\natexlab{}.
\newblock \showarticletitle{Apictorial Jigsaw Puzzles: The Computer Solution of
  a Problem in Pattern Recognition}.
\newblock \bibinfo{journal}{\emph{IEEE Transactions on Electronic Computers}}
  \bibinfo{volume}{EC-13}, \bibinfo{number}{2} (\bibinfo{year}{1964}),
  \bibinfo{pages}{118--127}.
\newblock
\href{https://doi.org/10.1109/PGEC.1964.263781}{doi:\nolinkurl{10.1109/PGEC.1964.263781}}


\bibitem[Funkhouser et~al\mbox{.}(2011)]%
        {Funkhouser2011}
\bibfield{author}{\bibinfo{person}{Thomas Funkhouser}, \bibinfo{person}{Hijung
  Shin}, \bibinfo{person}{Corey Toler-Franklin},
  \bibinfo{person}{Antonio~Garc{\'\i}a Casta{\~n}eda},
  \bibinfo{person}{Benedict Brown}, \bibinfo{person}{David Dobkin},
  \bibinfo{person}{Szymon Rusinkiewicz}, {and} \bibinfo{person}{Tim Weyrich}.}
  \bibinfo{year}{2011}\natexlab{}.
\newblock \showarticletitle{Learning How to Match Fresco Fragments}.
\newblock \bibinfo{journal}{\emph{Journal on Computing and Cultural Heritage}}
  \bibinfo{volume}{4}, \bibinfo{number}{2} (\bibinfo{year}{2011}),
  \bibinfo{pages}{7:1--7:13}.
\newblock
\href{https://doi.org/10.1145/2037820.2037824}{doi:\nolinkurl{10.1145/2037820.2037824}}


\bibitem[Gigilashvili et~al\mbox{.}(2024)]%
        {Gigilashvili2024Textile}
\bibfield{author}{\bibinfo{person}{Davit Gigilashvili},
  \bibinfo{person}{Casper~Fabian Gulbrandsen}, \bibinfo{person}{Ha~Thu Nguyen},
  \bibinfo{person}{Margrethe Havgar}, \bibinfo{person}{Marianne Vedeler}, {and}
  \bibinfo{person}{Jon~Yngve Hardeberg}.} \bibinfo{year}{2024}\natexlab{}.
\newblock \showarticletitle{Toward Solving a Puzzle of Fragmented Archeological
  Textiles}.
\newblock \bibinfo{journal}{\emph{Journal of Imaging Science and Technology}}
  \bibinfo{volume}{68}, \bibinfo{number}{4} (\bibinfo{year}{2024}),
  \bibinfo{pages}{1--16}.
\newblock
\href{https://doi.org/10.2352/J.ImagingSci.Technol.2024.68.4.040407}{doi:\nolinkurl{10.2352/J.ImagingSci.Technol.2024.68.4.040407}}


\bibitem[Gigilashvili et~al\mbox{.}(2023a)]%
        {Gigilashvili2023heritage}
\bibfield{author}{\bibinfo{person}{Davit Gigilashvili}, \bibinfo{person}{Hana
  Lukesova}, \bibinfo{person}{Casper~Fabian Gulbrandsen},
  \bibinfo{person}{Akash Harijan}, {and} \bibinfo{person}{Jon~Yngve
  Hardeberg}.} \bibinfo{year}{2023}\natexlab{a}.
\newblock \showarticletitle{Computational techniques for virtual reconstruction
  of fragmented archaeological textiles}.
\newblock \bibinfo{journal}{\emph{Heritage Science}} \bibinfo{volume}{11},
  \bibinfo{number}{259} (\bibinfo{year}{2023}).
\newblock


\bibitem[Gigilashvili et~al\mbox{.}(2023b)]%
        {Gigilashvili2023archiving}
\bibfield{author}{\bibinfo{person}{Davit Gigilashvili}, \bibinfo{person}{Ha~Thu
  Nguyen}, \bibinfo{person}{Casper~F. Gulbrandsen}, \bibinfo{person}{Margrethe
  Havgar}, \bibinfo{person}{Marianne Vedeler}, {and} \bibinfo{person}{Jon~Y.
  Hardeberg}.} \bibinfo{year}{2023}\natexlab{b}.
\newblock \showarticletitle{Texture-based Clustering of Archaeological Textile
  Images}. In \bibinfo{booktitle}{\emph{Proceedings of Archiving 2023}},
  Vol.~\bibinfo{volume}{20}. Society for Imaging Science and Technology,
  \bibinfo{pages}{139--142}.
\newblock
\href{https://doi.org/10.2352/issn.2168-3204.2023.20.1.29}{doi:\nolinkurl{10.2352/issn.2168-3204.2023.20.1.29}}


\bibitem[Goldberg et~al\mbox{.}(2004)]%
        {Goldberg2002}
\bibfield{author}{\bibinfo{person}{D Goldberg}, \bibinfo{person}{C Malon},
  {and} \bibinfo{person}{M Bern}.} \bibinfo{year}{2004}\natexlab{}.
\newblock \showarticletitle{A global approach to automatic solution of jigsaw
  puzzles}.
\newblock \bibinfo{journal}{\emph{Computational Geometry}}
  \bibinfo{volume}{28}, \bibinfo{number}{2-3} (\bibinfo{year}{2004}),
  \bibinfo{pages}{165--174}.
\newblock


\bibitem[Gulbrandsen(2023)]%
        {Gulbrandsen2023}
\bibfield{author}{\bibinfo{person}{Casper~Fabian Gulbrandsen}.}
  \bibinfo{year}{2023}\natexlab{}.
\newblock \emph{\bibinfo{title}{Reconstructing the Original: Machine Learning
  Puzzle Assembly for Archaeological Textile Reconstruction}}.
\newblock \bibinfo{thesistype}{Master's\ thesis}. \bibinfo{school}{Norwegian
  University of Science and Technology}.
\newblock


\bibitem[He et~al\mbox{.}(2016)]%
        {ResNet2016}
\bibfield{author}{\bibinfo{person}{Kaiming He}, \bibinfo{person}{Xiangyu
  Zhang}, \bibinfo{person}{Shaoqing Ren}, {and} \bibinfo{person}{Jian Sun}.}
  \bibinfo{year}{2016}\natexlab{}.
\newblock \showarticletitle{Deep Residual Learning for Image Recognition}. In
  \bibinfo{booktitle}{\emph{IEEE Conference on Computer Vision and Pattern
  Recognition (CVPR)}}. \bibinfo{pages}{770--778}.
\newblock
\href{https://doi.org/10.1109/CVPR.2016.90}{doi:\nolinkurl{10.1109/CVPR.2016.90}}


\bibitem[Ho et~al\mbox{.}(2020)]%
        {DDPM2020}
\bibfield{author}{\bibinfo{person}{Jonathan Ho}, \bibinfo{person}{Ajay~N.
  Jain}, {and} \bibinfo{person}{Pieter Abbeel}.}
  \bibinfo{year}{2020}\natexlab{}.
\newblock \showarticletitle{Denoising Diffusion Probabilistic Models}. In
  \bibinfo{booktitle}{\emph{Advances in Neural Information Processing
  Systems}}, Vol.~\bibinfo{volume}{33}.
\newblock


\bibitem[Karras et~al\mbox{.}(2018)]%
        {CelebAHQ2018}
\bibfield{author}{\bibinfo{person}{Tero Karras}, \bibinfo{person}{Timo Aila},
  \bibinfo{person}{Samuli Laine}, {and} \bibinfo{person}{Jaakko Lehtinen}.}
  \bibinfo{year}{2018}\natexlab{}.
\newblock \showarticletitle{Progressive Growing of {GANs} for Improved Quality,
  Stability, and Variation}. In \bibinfo{booktitle}{\emph{International
  Conference on Learning Representations (ICLR)}}.
\newblock
\newblock
\shownote{Introduces the CelebA-HQ dataset of 30,000 1024$\times$1024 images.}.


\bibitem[Kim et~al\mbox{.}(2025)]%
        {FCViT2025}
\bibfield{author}{\bibinfo{person}{Garam Kim}, \bibinfo{person}{Hyeonseong
  Cho}, {and} \bibinfo{person}{Hyoungsik Nam}.}
  \bibinfo{year}{2025}\natexlab{}.
\newblock \showarticletitle{Solving Jigsaw Puzzles by Predicting Fragment's
  Coordinate Based on Vision Transformer (FCViT)}.
\newblock \bibinfo{journal}{\emph{Expert Systems with Applications}}
  \bibinfo{volume}{272} (\bibinfo{year}{2025}), \bibinfo{pages}{126776}.
\newblock
\href{https://doi.org/10.1016/j.eswa.2025.126776}{doi:\nolinkurl{10.1016/j.eswa.2025.126776}}


\bibitem[LanceDB(2024)]%
        {LanceDB}
LanceDB \bibinfo{year}{2024}\natexlab{}.
\newblock \bibinfo{title}{{LanceDB}: Developer-friendly, embedded retrieval
  engine for multimodal {AI}}.
\newblock \bibinfo{howpublished}{\url{https://github.com/lancedb/lancedb}}.
\newblock
\newblock
\shownote{Open-source software, Apache-2.0 license}.


\bibitem[Liu et~al\mbox{.}(2024)]%
        {JPDVT2024}
\bibfield{author}{\bibinfo{person}{Jinyang Liu}, \bibinfo{person}{Wondmgezahu
  Teshome}, \bibinfo{person}{Sandesh Ghimire}, \bibinfo{person}{Mario Sznaier},
  {and} \bibinfo{person}{Octavia Camps}.} \bibinfo{year}{2024}\natexlab{}.
\newblock \showarticletitle{Solving Masked Jigsaw Puzzles with Diffusion Vision
  Transformers}. In \bibinfo{booktitle}{\emph{Proceedings of the IEEE/CVF
  Conference on Computer Vision and Pattern Recognition (CVPR)}}.
\newblock


\bibitem[Loshchilov and Hutter(2019)]%
        {AdamW2019}
\bibfield{author}{\bibinfo{person}{Ilya Loshchilov} {and}
  \bibinfo{person}{Frank Hutter}.} \bibinfo{year}{2019}\natexlab{}.
\newblock \showarticletitle{Decoupled Weight Decay Regularization}. In
  \bibinfo{booktitle}{\emph{International Conference on Learning
  Representations (ICLR)}}.
\newblock


\bibitem[McCloskey and Cohen(1989)]%
        {McCloskey1989}
\bibfield{author}{\bibinfo{person}{Michael McCloskey} {and}
  \bibinfo{person}{Neal~J. Cohen}.} \bibinfo{year}{1989}\natexlab{}.
\newblock \showarticletitle{Catastrophic Interference in Connectionist
  Networks: The Sequential Learning Problem}.
\newblock In \bibinfo{booktitle}{\emph{Psychology of Learning and Motivation}},
  \bibfield{editor}{\bibinfo{person}{Gordon~H. Bower}} (Ed.).
  Vol.~\bibinfo{volume}{24}. \bibinfo{publisher}{Academic Press},
  \bibinfo{pages}{109--165}.
\newblock
\href{https://doi.org/10.1016/S0079-7421(08)60536-8}{doi:\nolinkurl{10.1016/S0079-7421(08)60536-8}}


\bibitem[McInnes et~al\mbox{.}(2018)]%
        {mcinnes2018umap}
\bibfield{author}{\bibinfo{person}{Leland McInnes}, \bibinfo{person}{John
  Healy}, {and} \bibinfo{person}{James Melville}.}
  \bibinfo{year}{2018}\natexlab{}.
\newblock \showarticletitle{UMAP: Uniform Manifold Approximation and Projection
  for Dimension Reduction}.
\newblock \bibinfo{journal}{\emph{arXiv preprint arXiv:1802.03426}}
  (\bibinfo{year}{2018}).
\newblock


\bibitem[Moore and Corso(2020)]%
        {moore2020fiftyone}
\bibfield{author}{\bibinfo{person}{B.~E. Moore} {and} \bibinfo{person}{J.~J.
  Corso}.} \bibinfo{year}{2020}\natexlab{}.
\newblock \showarticletitle{FiftyOne}.
\newblock \bibinfo{journal}{\emph{GitHub}} (\bibinfo{year}{2020}).
\newblock
\newblock
\shownote{https://github.com/voxel51/fiftyone}.


\bibitem[Noroozi and Favaro(2016)]%
        {Noroozi2016}
\bibfield{author}{\bibinfo{person}{Mehdi Noroozi} {and} \bibinfo{person}{Paolo
  Favaro}.} \bibinfo{year}{2016}\natexlab{}.
\newblock \showarticletitle{Unsupervised Learning of Visual Representations by
  Solving Jigsaw Puzzles}. In \bibinfo{booktitle}{\emph{Computer Vision -- ECCV
  2016}} \emph{(\bibinfo{series}{Lecture Notes in Computer Science},
  Vol.~\bibinfo{volume}{9910})}. \bibinfo{pages}{69--84}.
\newblock
\href{https://doi.org/10.1007/978-3-319-46466-4_5}{doi:\nolinkurl{10.1007/978-3-319-46466-4_5}}


\bibitem[Paumard et~al\mbox{.}(2018a)]%
        {Paumard2018shortest}
\bibfield{author}{\bibinfo{person}{Marie-Morgane Paumard},
  \bibinfo{person}{David Picard}, {and} \bibinfo{person}{Hedi Tabia}.}
  \bibinfo{year}{2018}\natexlab{a}.
\newblock \showarticletitle{Image Reassembly Combining Deep Learning and
  Shortest Path Problem}. In \bibinfo{booktitle}{\emph{Computer Vision -- ECCV
  2018}} \emph{(\bibinfo{series}{LNCS}, Vol.~\bibinfo{volume}{11210})}.
  \bibinfo{pages}{155--169}.
\newblock
\href{https://doi.org/10.1007/978-3-030-01231-1\_10}{doi:\nolinkurl{10.1007/978-3-030-01231-1\_10}}


\bibitem[Paumard et~al\mbox{.}(2018b)]%
        {Paumard2018}
\bibfield{author}{\bibinfo{person}{Marie-Morgane Paumard},
  \bibinfo{person}{David Picard}, {and} \bibinfo{person}{Hedi Tabia}.}
  \bibinfo{year}{2018}\natexlab{b}.
\newblock \showarticletitle{Jigsaw Puzzle Solving Using Local Feature
  Co-Occurrences in Deep Neural Networks}.
\newblock \bibinfo{journal}{\emph{arXiv preprint arXiv:1807.03155}}
  (\bibinfo{year}{2018}).
\newblock
\href{https://doi.org/10.48550/arXiv.1807.03155}{doi:\nolinkurl{10.48550/arXiv.1807.03155}}


\bibitem[Pomeranz et~al\mbox{.}(2011)]%
        {Pomeranz_CVPR_2011}
\bibfield{author}{\bibinfo{person}{Dolev Pomeranz}, \bibinfo{person}{Michal
  Shemesh}, {and} \bibinfo{person}{Ohad Ben-Shahar}.}
  \bibinfo{year}{2011}\natexlab{}.
\newblock \showarticletitle{A Fully Automated Greedy Square Jigsaw Puzzle
  Solver}. In \bibinfo{booktitle}{\emph{Proc.\ IEEE Conf.\ on Computer Vision
  and Pattern Recognition (CVPR)}}. \bibinfo{pages}{9--16}.
\newblock
\href{https://doi.org/10.1109/CVPR.2011.5995331}{doi:\nolinkurl{10.1109/CVPR.2011.5995331}}


\bibitem[Radford et~al\mbox{.}(2021)]%
        {radford2021clip}
\bibfield{author}{\bibinfo{person}{Alec Radford}, \bibinfo{person}{Jong~Wook
  Kim}, \bibinfo{person}{Chris Hallacy}, \bibinfo{person}{Aditya Ramesh},
  \bibinfo{person}{Gabriel Goh}, \bibinfo{person}{Sandhini Agarwal},
  \bibinfo{person}{Girish Sastry}, \bibinfo{person}{Amanda Askell},
  \bibinfo{person}{Pamela Mishkin}, \bibinfo{person}{Jack Clark},
  \bibinfo{person}{Gretchen Krueger}, {and} \bibinfo{person}{Ilya Sutskever}.}
  \bibinfo{year}{2021}\natexlab{}.
\newblock \showarticletitle{{CLIP}: Learning Transferable Visual Models From
  Natural Language Supervision}.
\newblock \bibinfo{journal}{\emph{arXiv preprint arXiv:2103.00020}}
  (\bibinfo{year}{2021}).
\newblock


\bibitem[Scarpellini et~al\mbox{.}(2024)]%
        {DiffAssemble2024}
\bibfield{author}{\bibinfo{person}{Gianluca Scarpellini},
  \bibinfo{person}{Stefano Fiorini}, \bibinfo{person}{Francesco Giuliari},
  \bibinfo{person}{Pietro Morerio}, {and} \bibinfo{person}{Alessio~Del Bue}.}
  \bibinfo{year}{2024}\natexlab{}.
\newblock \showarticletitle{DiffAssemble: A Unified Graph-Diffusion Model for
  2D and 3D Reassembly}. In \bibinfo{booktitle}{\emph{Proceedings of the
  IEEE/CVF Conference on Computer Vision and Pattern Recognition (CVPR)}}.
  \bibinfo{pages}{28098--28108}.
\newblock


\bibitem[Sholomon et~al\mbox{.}(2013)]%
        {Sholomon2013}
\bibfield{author}{\bibinfo{person}{Dror Sholomon}, \bibinfo{person}{Omid
  David}, {and} \bibinfo{person}{Nathan~S. Netanyahu}.}
  \bibinfo{year}{2013}\natexlab{}.
\newblock \showarticletitle{A Genetic Algorithm-Based Solver for Very Large
  Jigsaw Puzzles}. In \bibinfo{booktitle}{\emph{Proceedings of the IEEE
  Conference on Computer Vision and Pattern Recognition (CVPR)}}.
  \bibinfo{pages}{1767--1774}.
\newblock
\href{https://doi.org/10.1109/CVPR.2013.231}{doi:\nolinkurl{10.1109/CVPR.2013.231}}


\bibitem[Song et~al\mbox{.}(2021)]%
        {DDIM2021}
\bibfield{author}{\bibinfo{person}{Jiaming Song}, \bibinfo{person}{Chenlin
  Meng}, {and} \bibinfo{person}{Stefano Ermon}.}
  \bibinfo{year}{2021}\natexlab{}.
\newblock \showarticletitle{Denoising Diffusion Implicit Models}. In
  \bibinfo{booktitle}{\emph{International Conference on Learning
  Representations (ICLR)}}.
\newblock


\bibitem[Song et~al\mbox{.}(2023)]%
        {JPwLEG3}
\bibfield{author}{\bibinfo{person}{Xingke Song}, \bibinfo{person}{Jiahuan Jin},
  \bibinfo{person}{Chenglin Yao}, \bibinfo{person}{Shihe Wang},
  \bibinfo{person}{Jianfeng Ren}, {and} \bibinfo{person}{Ruibin Bai}.}
  \bibinfo{year}{2023}\natexlab{}.
\newblock \showarticletitle{Siamese-Discriminant Deep Reinforcement Learning
  for Solving Jigsaw Puzzles with Large Eroded Gaps}. In
  \bibinfo{booktitle}{\emph{Proceedings of the AAAI Conference on Artificial
  Intelligence}}, Vol.~\bibinfo{volume}{37}. \bibinfo{pages}{2303--2311}.
\newblock
\href{https://doi.org/10.1609/aaai.v37i2.25325}{doi:\nolinkurl{10.1609/aaai.v37i2.25325}}


\bibitem[Tan and Le(2019)]%
        {EfficientNet2019}
\bibfield{author}{\bibinfo{person}{Mingxing Tan} {and} \bibinfo{person}{Quoc
  Le}.} \bibinfo{year}{2019}\natexlab{}.
\newblock \showarticletitle{{EfficientNet}: Rethinking Model Scaling for
  Convolutional Neural Networks}. In \bibinfo{booktitle}{\emph{Proceedings of
  the 36th International Conference on Machine Learning}}
  \emph{(\bibinfo{series}{Proceedings of Machine Learning Research},
  Vol.~\bibinfo{volume}{97})}. \bibinfo{pages}{6105--6114}.
\newblock


\bibitem[{The Metropolitan Museum of Art}(2024)]%
        {MET_OpenAccess_2024}
\bibfield{author}{\bibinfo{person}{{The Metropolitan Museum of Art}}.}
  \bibinfo{year}{2024}\natexlab{}.
\newblock \bibinfo{title}{Open Access Artworks Dataset}.
\newblock
  \bibinfo{howpublished}{\url{https://github.com/metmuseum/openaccess}}.
\newblock
\newblock
\shownote{CC0 public-domain images and metadata. Accessed: 2025-01-15}.


\bibitem[Tsesmelis et~al\mbox{.}(2024)]%
        {Tsesmelis2024}
\bibfield{author}{\bibinfo{person}{Theodore Tsesmelis}, \bibinfo{person}{Luca
  Palmieri}, \bibinfo{person}{Marina Khoroshiltseva}, \bibinfo{person}{Marcello
  Pelillo}, \bibinfo{person}{Alessio~Del Bue}, {et~al\mbox{.}}}
  \bibinfo{year}{2024}\natexlab{}.
\newblock \showarticletitle{Re-assembling the Past: The RePAIR Dataset and
  Benchmark for Real-World 2D and 3D Puzzle Solving}. In
  \bibinfo{booktitle}{\emph{Proceedings of the 38th International Conference on
  Neural Information Processing Systems}}. \bibinfo{pages}{30076--30105}.
\newblock


\bibitem[Vaswani et~al\mbox{.}(2017)]%
        {Vaswani2017}
\bibfield{author}{\bibinfo{person}{Ashish Vaswani}, \bibinfo{person}{Noam
  Shazeer}, \bibinfo{person}{Niki Parmar}, \bibinfo{person}{Jakob Uszkoreit},
  \bibinfo{person}{Llion Jones}, \bibinfo{person}{Aidan~N. Gomez},
  \bibinfo{person}{{\L}ukasz Kaiser}, {and} \bibinfo{person}{Illia
  Polosukhin}.} \bibinfo{year}{2017}\natexlab{}.
\newblock \showarticletitle{Attention Is All You Need}. In
  \bibinfo{booktitle}{\emph{Advances in Neural Information Processing
  Systems}}, Vol.~\bibinfo{volume}{30}. \bibinfo{pages}{5998--6008}.
\newblock


\bibitem[Vedeler(2019)]%
        {Vedeler2019}
\bibfield{author}{\bibinfo{person}{Marianne Vedeler}.}
  \bibinfo{year}{2019}\natexlab{}.
\newblock \bibinfo{booktitle}{\emph{The Oseberg Tapestries}}.
\newblock \bibinfo{publisher}{Scandinavian Academic Press, Oslo, Norway}.
\newblock


\bibitem[Wang et~al\mbox{.}(2025)]%
        {ASUKA2025}
\bibfield{author}{\bibinfo{person}{Yikai Wang}, \bibinfo{person}{Chenjie Cao},
  \bibinfo{person}{Junqiu Yu}, \bibinfo{person}{Ke Fan},
  \bibinfo{person}{Xiangyang Xue}, {and} \bibinfo{person}{Yanwei Fu}.}
  \bibinfo{year}{2025}\natexlab{}.
\newblock \showarticletitle{Towards Enhanced Image Inpainting: Mitigating
  Unwanted Object Insertion and Preserving Color Consistency}. In
  \bibinfo{booktitle}{\emph{Proceedings of the IEEE/CVF Conference on Computer
  Vision and Pattern Recognition (CVPR)}}. \bibinfo{pages}{23237--23248}.
\newblock


\bibitem[Zhong et~al\mbox{.}(2023)]%
        {TextileNet2020}
\bibfield{author}{\bibinfo{person}{Shu Zhong}, \bibinfo{person}{Miriam Ribul},
  \bibinfo{person}{Youngjun Cho}, {and} \bibinfo{person}{Marianna Obrist}.}
  \bibinfo{year}{2023}\natexlab{}.
\newblock \showarticletitle{TextileNet: A Material Taxonomy-based Fashion
  Textile Dataset}.
\newblock \bibinfo{journal}{\emph{arXiv preprint arXiv:2301.06160}}
  (\bibinfo{year}{2023}).
\newblock


\end{thebibliography}

\end{document}